\documentclass[10]{article}

\usepackage[rgb]{xcolor}
\usepackage{pgf}
\makeatletter
\@namedef{algorithm.sty}{}
\makeatother
\makeatletter
\@namedef{algorithmic.sty}{}
\makeatother
\usepackage[accepted]{icml2019}
\usepackage[utf8]{inputenc}
\usepackage{pgfplots}
\usepackage{amsfonts}
\usepackage{amsmath}
\usepackage{amstext}
\usepackage{nidanfloat}  
\usepackage{xspace}
\usepackage{subfigure}
\usepackage{multirow}
\usepgfplotslibrary{groupplots}
\pgfplotsset{compat=newest}
\usepackage{shellesc}
\usepackage{booktabs}
\usepackage[english]{babel}
\usepackage[]{hyperref}
\usepackage{url}
\usepackage{placeins}
\usepackage[]{caption}
\usepackage[linesnumbered,ruled,noend]{algorithm2e}
\usepackage[font=small,labelfont=bf,tableposition=top]{caption}
\DeclareCaptionLabelFormat{andtable}{#1~#2  \&  \tablename~\thetable}

\addto\extrasenglish{%

}

\usetikzlibrary{external}
\usepackage{titlesec}

\begin{document}

\definecolor{mygreen}{rgb}{0.2, 0.7, 0.2}
\definecolor{myorange}{rgb}{0.9, 0.5, 0.0}

\definecolor{blue}{HTML}{5DA5DA}
\definecolor{orange}{HTML}{FAA43A} 
\definecolor{green}{HTML}{60BD68} 
\definecolor{pink}{HTML}{F17CB0} 
\definecolor{brown}{HTML}{B2912F} 
\definecolor{purple}{HTML}{B276B2} 
\definecolor{yellow}{HTML}{DECF3F} 
\definecolor{red}{HTML}{F15854} 
\definecolor{gray}{HTML}{4D4D4D} 

\newcommand\mathdefault[1]{#1}

\newcommand\noteMF[1]{\textcolor{red}{MF - #1}}
\newcommand\noteKC[1]{\textcolor{blue}{KC - #1}}
\newcommand\notePM[1]{\textcolor{orange}{PM - #1}}
\newcommand\noteSR[1]{\textcolor{green}{SR - #1}}

\newcommand{\nobs}{n} 
\newcommand{\R}{\mathbb{R}}
\newcommand{\N}{\mathcal{N}}
\newcommand{\B}{\mathcal{B}}
\newcommand{\Z}{\mathbb{Z}}
\newcommand{\F}{\mathcal{F}}
\newcommand{\D}{\mathcal{D}}
\newcommand{\X}{\mathcal{X}}
\newcommand{\Y}{\mathcal{Y}}
\newcommand{\I}{\mathcal{I}}
\newcommand{\LL}{\mathcal{L}}
\newcommand{\uu}{\mathbf{u}}
\newcommand{\ee}{\mathbf{e}}

\newcommand{\T}{\top}
\newcommand{\E}{\mathrm{E}}
\newcommand{\const}{\mathrm{const.}}
\newcommand{\diag}{\mathrm{diag}}
\newcommand{\Tr}{\mathrm{Tr}}

\newcommand{\norm}{\mathcal{N}}

\newcommand{\avect}{\mathbf{a}}
\newcommand{\dvect}{\mathbf{d}}
\newcommand{\fvect}{\mathbf{f}}
\newcommand{\gvect}{\mathbf{g}}
\newcommand{\Ivect}{\mathbf{I}}
\newcommand{\Kvect}{\mathbf{K}}
\newcommand{\hvect}{\mathbf{h}}
\newcommand{\Lvect}{\mathbf{L}}
\newcommand{\mvect}{\mathbf{m}}
\newcommand{\pvect}{\mathbf{p}}
\newcommand{\svect}{\mathbf{s}}
\newcommand{\Svect}{\mathbf{S}}
\newcommand{\uvect}{\mathbf{u}}
\newcommand{\Uvect}{\mathbf{U}}
\newcommand{\vvect}{\mathbf{v}}
\newcommand{\Vvect}{\mathbf{V}}
\newcommand{\zvect}{\mathbf{z}}
\newcommand{\xvect}{\mathbf{x}}
\newcommand{\Xvect}{\mathbf{X}}
\newcommand{\yvect}{\mathbf{y}}
\newcommand{\Yvect}{\mathbf{Y}}
\newcommand{\wvect}{\mathbf{w}}
\newcommand{\Wvect}{\mathbf{W}}
\newcommand{\tvect}{\mathbf{t}}
\newcommand{\zerovect}{\mathbf{0}}
\newcommand{\onesvect}{\mathbf{1}}

\newcommand{\Dobs}{D_\mathrm{obs}}
\newcommand{\Dlat}{D_\mathrm{lat}}
\newcommand{\Din}{D_\mathrm{in}}
\newcommand{\Dout}{D_\mathrm{out}}

\newcommand{\betavect}{\boldsymbol{\beta}}
\newcommand{\thetavect}{\boldsymbol{\theta}}
\newcommand{\Thetavect}{\boldsymbol{\Theta}}
\newcommand{\psivect}{\boldsymbol{\psi}}
\newcommand{\Psivect}{\boldsymbol{\Psi}}
\newcommand{\Phivect}{\boldsymbol{\Phi}}
\newcommand{\Pivect}{\boldsymbol{\Pi}}
\newcommand{\etavect}{\boldsymbol{\eta}}
\newcommand{\rhovect}{\boldsymbol{\rho}}
\newcommand{\tauvect}{\boldsymbol{\tau}}
\newcommand{\nuvect}{\boldsymbol{\nu}}
\newcommand{\muvect}{\boldsymbol{\mu}}
\newcommand{\omegavect}{\boldsymbol{\omega}}
\newcommand{\Omegavect}{\boldsymbol{\Omega}}
\newcommand{\sigmavect}{\boldsymbol{\sigma}}
\newcommand{\zetavect}{\boldsymbol{\zeta}}
\newcommand{\varepsilonvect}{\boldsymbol{\epsilon}}
\newcommand{\deltavect}{\boldsymbol{\delta}}

\newcommand{\bigO}{\mathcal{O}}

\newcommand{\name}[1]{{\textsc{#1}}\xspace}

\newcommand{\mcmc}{\name{mcmc}}
\newcommand{\vi}{\name{vi}}

\newcommand{\autogp}{\textsc{a}uto\textsc{gp}\xspace}
\newcommand{\gpflow}{\textsc{gpf}low\xspace}
\newcommand{\tensorflow}{\textsc{T}ensor\textsc{F}low\xspace}
\newcommand{\pytorch}{\name{pytorch}}

\newcommand{\arccosine}{\name{arc-cosine}}
\newcommand{\dbnthree}{\name{DBN}}
\newcommand{\svdkl}{\name{SV-DKL}}
\newcommand{\dgprbf}{\textsc{dgp}-\textsc{rbf}\xspace}
\newcommand{\dgparc}{\textsc{dgp}-\textsc{arc}\xspace}
\newcommand{\dgpep}{\textsc{dgp}-\textsc{ep}\xspace}
\newcommand{\dgpvar}{\textsc{dgp}-\textsc{var}\xspace}

\newcommand{\gplvm}{\name{gplvm}}
\newcommand{\dgplvm}{\name{dgplvm}}
\newcommand{\gp}{\name{gp}}
\newcommand{\gps}{\textsc{gp}s\xspace}
\newcommand{\dgp}{\name{dgp}}
\newcommand{\dgps}{\textsc{dgp}s\xspace}
\newcommand{\dnn}{\name{dnn}}
\newcommand{\dnns}{\textsc{dnn}s\xspace}
\newcommand{\svm}{\name{svm}}
\newcommand{\svms}{\textsc{svm}s\xspace}
\newcommand{\vargp}{\textsc{var}-\textsc{gp}\xspace}

\newcommand{\cnn}{\name{cnn}}
\newcommand{\bnn}{\name{bnn}}
\newcommand{\bcnn}{\name{bcnn}}
\newcommand{\cnns}{\textsc{cnn}s\xspace}
\newcommand{\mcd}{\name{mcd}}

\newcommand{\ard}{\name{ard}}
\newcommand{\softmax}{\name{softmax}}

\newcommand{\relu}{{\textsc{r}}e\name{lu}}

\newcommand{\arc}{\name{arc}}
\newcommand{\rbf}{\name{rbf}}

\newcommand{\adam}{\name{adam}}

\newcommand{\nll}{\name{nll}}
\newcommand{\mnll}{\name{mnll}}
\newcommand{\ece}{\name{ece}}
\newcommand{\errate}{\name{error rate}}
\newcommand{\err}{\name{err}}
\newcommand{\rmse}{\name{rmse}}
\newcommand{\nelbo}{\name{nelbo}}

\newcommand{\convnets}{\textsc{convnet}s}

\newcommand{\cifar}{\name{cifar10}}
\newcommand{\mnisteight}{\textsc{mnist}8\textsc{m}\xspace}
\newcommand{\protein}{\name{protein}}
\newcommand{\powerplant}{\name{powerplant}}
\newcommand{\spam}{\name{spam}}
\newcommand{\credit}{\name{credit}}
\newcommand{\eeg}{\name{eeg}}
\newcommand{\mnist}{\name{mnist}}
\newcommand{\notmnist}{\name{not-mnist}}
\newcommand{\boston}{\name{boston}}
\newcommand{\concrete}{\name{concrete}}

\newcommand{\cifart}{\name{cifar10}}
\newcommand{\cifarh}{\name{cifar100}}

\newcommand{\iblm}{\name{i-blm}}
\newcommand{\blm}{\name{blm}}
\newcommand{\blms}{\textsc{blm}s\xspace}
\newcommand{\uninformative}{\name{uninformative}}
\newcommand{\heuristic}{\name{heuristic}}
\newcommand{\orthogonal}{\name{orthogonal}}
\newcommand{\lsuv}{\name{lsuv}}
\newcommand{\xavier}{\name{xavier}}

\newcommand{\svi}{\name{svi}}

\newcommand{\lenet}{\name{LeNet-5}}
\newcommand{\resnet}{\name{resnet}}

\newcommand{\alexnet}{\name{AlexNet}}
\newcommand{\vgg}{\name{vgg16}}

\newcommand{\Exp}{\mathbb{E}}

\newcommand{\KL}{\name{kl}}

\newcommand{\dkl}[2]{{\KL}\left(#1\vert\vert#2\right)}

\newcommand{\expnot}[2]{{#1}\mathrm{e}{\text{-}#2}}

\twocolumn[
\icmltitle{Good Initializations of Variational Bayes for Deep Models}
\icmltitlerunning{Good Initializations of Variational Bayes for Deep Models}

\begin{icmlauthorlist}
\icmlauthor{Simone Rossi}{eurecom}
\icmlauthor{Pietro Michiardi}{eurecom}
\icmlauthor{Maurizio Filippone}{eurecom}
\end{icmlauthorlist}

\icmlaffiliation{eurecom}{Department of Data Science, EURECOM, France}

\icmlcorrespondingauthor{Simone Rossi}{simone.rossi@eurecom.fr}
\icmlcorrespondingauthor{Pietro Michiardi}{pietro.michiardi@eurecom.fr}
\icmlcorrespondingauthor{Maurizio Filippone}{maurizio.filippone@eurecom.fr}
\icmlkeywords{Machine Learning, ICML}

\vskip 0.3in
]



\printAffiliationsAndNotice{}  

\begin{abstract}
    Stochastic variational inference is an established way to carry out approximate Bayesian inference for deep models.
    While there have been effective proposals for good initializations for loss minimization in deep learning, far less attention has been devoted to the issue of initialization of stochastic variational inference.
    We address this by proposing a novel layer-wise initialization strategy based on Bayesian linear models.
    The proposed method is extensively validated on regression and classification tasks, including Bayesian \textsc{DeepNets} and \textsc{ConvNets}, showing faster and better convergence compared to alternatives inspired by the literature on initializations for loss minimization.
\end{abstract}

\newlength\figureheight
\newlength\figurewidth

\section{Introduction}
\label{sec:intro}
Deep Neural Networks (\dnns) and Convolutional Neural Networks (\cnns) have become the preferred choice to tackle various supervised learning problems, such as regression and classification, due to their ability to model complex problems and the mature development of regularization techniques to control overfitting \citep{LeCun15, Srivastava14}.
There has been a recent surge of interest in the issues associated with their overconfidence in predictions, and proposals to mitigate these \citep{Guo17,Kendall17,Lakshminarayanan17}. 
Bayesian techniques offer a natural framework to deal with such issues, but they are characterized by computational intractability \citep{Bishop06,Ghahramani15}. 

A popular way to recover tractability is to use variational inference \citep{Jordan99}.
In variational inference, an approximate posterior distribution is introduced and its parameters are adapted by optimizing a variational objective, which is a lower bound to the marginal likelihood.
The variational objective can be written as the sum of an expectation of the log-likelihood under the approximate posterior and a regularization term which is the negative Kullback-Leibler (\KL) divergence between the approximating distribution and the prior over the parameters.
Stochastic Variational Inference (\svi) offers a practical way to carry out stochastic optimization of the variational objective. 
In \svi, stochasticity is introduced with a doubly stochastic approximation of the expectation term, which is unbiasedly approximated using Monte Carlo and by selecting a subset of the training points (mini-batching) \citep{Graves11,Kingma14}.

\begin{figure}[t]
    \centering
    \hspace{-0.01\textwidth}
    \setlength\figureheight{.2\textwidth}
    \setlength\figurewidth{.28\textwidth} 
    \subfigure{
        \tiny
        \ifdefined\compilefigures
            \tikzsetnextfilename{demo-convergence-uninf}
            \input{../fig/demo-convergence-exp/demo-after-uninf}
        \else
            \includegraphics{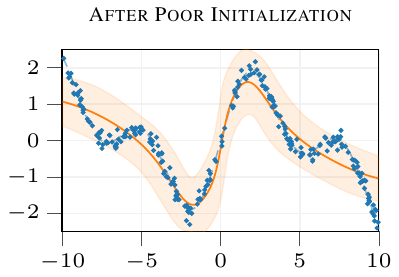}
        \fi
    }
    \hspace{-0.035\textwidth}
    \subfigure{
        \tiny
        \ifdefined\compilefigures
            \tikzsetnextfilename{demo-convergence-blm}
            \input{../fig/demo-convergence-exp/demo-after-blm}
        \else
            \includegraphics{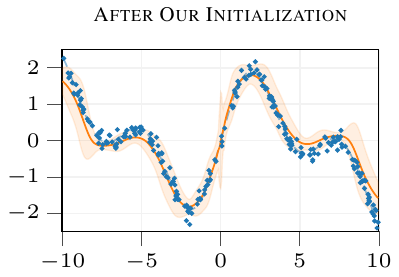}
        \fi
    }
    \caption{Due to poor initialization (\textbf{left}) \svi fails to converge even after $600+$ epochs ($\rmse = 0.613$, $\mnll = 29.4$) while with our \iblm (\textbf{right}) \svi easily recovers the function after few epochs ($\rmse = 0.315$, $\mnll = -5.8$). The architecture has three hidden layers with $500$ neurons each, and uses the \textsc{tanh} activation function. }
    \label{fig:demo-convergence}

\end{figure}

While \svi is an attractive and practical way to perform approximate inference for \dnns, there are limitations. 
For example, the form of the approximating distribution can be too simple to accurately approximate complex posterior distributions \citep{Ha16,Ranganath15,Rezende2015}. 
Furthermore, \svi increases the number of optimization parameters compared to optimizing model parameters through, e.g., loss minimization; for example, a fully factorized Gaussian posterior over model parameters doubles the number of parameters in the optimization compared to loss minimization. 
This has motivated research into other ways to perform approximate Bayesian inference for \dnns by establishing connections between variational inference and dropout \citep{Gal16, Gal16b, Gal17}. 

The development of a theory to fully understand the optimization landscape of \dnns and \cnns is still in its infancy \citep{Dziugaite17} and most works have focused on the practical aspects characterizing the optimization of their parameters \citep{Duchi11, Kingma2015b,Srivastava14}. 
If this lack of theory is apparent for optimization of model parameters, this is even more so for the understanding of the optimization landscape of the objective in variational inference, where variational parameters enter in a nontrivial way in the objective \citep{Graves11,Rezende14}. 
Initialization plays a huge role in the convergence of \svi; the illustrative example in \autoref{fig:demo-convergence} 
shows how a poor initialization can prevent \svi to converge to good solutions in short amount of time even for simple problems. 
The problem is even more severe for complex architectures, such as the ones that we discuss in the experiments; for example, \svi systematically converges to trivial solutions (posterior equal to the prior) when applied to \cnns, due to the difficulty in initializing variational parameters sensibly.

In this work, we focus on this issue affecting \svi for \dnns and \cnns.
While there is an established literature on ways to initialize model parameters of \dnns when minimizing its loss \citep{Glorot2010, Saxe2013, Mishkin2015}, to the best of our knowledge, there is no study that systematically tackle this issue for \svi for Bayesian \dnns and \cnns.
Inspired by the literature on residual networks \citep{He16} and greedy initialization of \dnns \citep{Bengio2006,Mishkin2015}, we propose Iterative Bayesian Linear Modeling (\iblm), which is an initialization strategy for \svi grounded on Bayesian linear modeling.
Iterating from the first layer, \iblm initializes the posteriors at layer $l$ by learning Bayesian linear models which regress from the input, propagated up to layer $l$, to the labels. 

We show how \iblm can be applied in a scalable way and without considerable overhead to regression and classification problems, and how it can be applied to initialize \svi not only for \dnns but also for \cnns. 
Through a series of experiments, we demonstrate that \iblm leads to faster convergence compared to other initalizations inspired by the work on loss minimization for \dnns. 
Furthermore, we show that \iblm makes it possible for \svi with a Gaussian approximation applied to \cnns to compete with Monte Carlo Dropout (\mcd; \citet{Gal16b}) and noisy natural gradients (\textsc{noisy-kfac}; \citet{Zhang2018}), which are state-of-art methods to perform approximate inference for \cnns. 
In all, thanks to the proposed initialization, we make it possible to reconsider Gaussian \svi for \dnns and \cnns as a valid competitor to \mcd and \textsc{noisy-kfac}, as well as highlight the limitations of \svi with a Gaussian posterior in applications involving \cnns.

In summary, in this work we make the following contributions: (1) we propose a novel way to initialize \svi for \dnns based on Bayesian linear models; (2) we show how this can be done for regression and classification; (3) we show how to apply our strategy to \cnns; (4) we empirically demonstrate that our proposal allows us to achieve performance superior to other initializations of \svi inspired by the literature on loss minimization; (5) for the first time, we achieve state-of-the-art performance with Gaussian \svi for large-scale \cnns.


\section{Related Work}
\label{ssec:related-works} 
The problem of initialization of weights and biases in \dnns for gradient-based loss minimization has been extensively tackled in the literature since early breakthroughs in the field \citep{Rumelhart1986,Baldi1989}.
\citet{LeCun1998b} is one of the seminal papers discussing practical tricks to achieve an efficient loss minimization through back-propagation. 

More recently, \citet{Bengio2006} propose a greedy layer-wise unsupervised pre-training that proved to help optimization and generalization. 
A justification can be found in \citet{Erhan2010}, where the authors show that pre-training can act as regularization; by initializing the parameters in a region corresponding to a better basin of attraction for the optimization procedure, the model can reach a better local minimum and increase its generalization capabilities. 
\citet{Glorot2010} propose a simple way to estimate the variance for random initialization of weights that makes it possible to avoid saturation both in forward and back-propagation steps. 
Another possible strategy can be found in \citet{Saxe2013}, that investigate the dynamics of gradient descend optimization, and propose an initialization based on random orthogonal initial conditions. 
This algorithm takes a weight matrix filled with Gaussian noise, decomposes it to orthonormal basis using a singular value decomposition and replaces the weights with one of the components. 
Building on this work, \citet{Mishkin2015} propose a data-driven weight initialization by scaling the orthonormal matrix of weights to make the variance of the output as close to one as possible.

Variational inference addresses the problem of intractable Bayesian inference by reinterpreting inference as an optimization problem. 
Its origins can be tracked back to early works in \citet{MacKay1992a, Hinton1993, Neal1997}. 
More recently, \citet{Graves11} proposes a practical way to carry out variational inference using stochastic optimization \citep{Duchi11, Zeiler2012, Sutskever2013, Kingma2015b}. 
\citet{Kingma14} propose a reparameterization trick that allows for the optimization of the variational lower bound through automatic differentiation. 
To decrease the variance of stochastic gradients, which impacts convergence speed, this work is extended using the so-called {\em local reparameterization trick}, where the sampling from the approximate posterior over model parameters is replaced by the sampling from the resulting distribution over the \dnn units \citep{Kingma2015}.

In the direction of finding richer posterior families for variational inference, we mention the works on Normalizing Flows \citep{Rezende2015, Kingma2016, Louizos2017,Huang2018}. 
Alternatives can be found in Stein variational inference \citep{Liu2016}, quasi-Monte Carlo variational inference \citep{Buchholz2018}, variational boosting \citep{Miller2016}, noisy natural gradients \citep{Zhang2018} and matrix Gaussian posterior \citep{Louizos2016b}.

To the best of our knowledge, there is no study that either empirically or theoretically addresses the problem of initialization of parameters for \svi; we could only find a mention of this in \citet{Krishnan2018} for variational autoencoders.
We aim to fill this gap by proposing a novel way to initialize parameters in \svi for probabilistic deep models.

\section{Preliminaries}
\label{sec:preliminaries}
In this section we introduce some background material on Bayesian \dnns and \svi.

\paragraph{Bayesian Deep Neural Networks} 
Bayesian \dnns are statistical models whose parameters (weights and biases) are assigned a prior distribution and inferred using Bayesian inference techniques.
Bayesian \dnns inherit the modeling capacity of \dnns while allowing for quantification of uncertainty in model parameters and predictions.
Considering an input $\xvect\in\R^{\Din}$ and a corresponding output $\yvect\in\R^{\Dout}$, the relation between inputs and outpus can be seen as a composition of nonlinear vector-valued functions $\fvect^{(l)}$ for each hidden layer $(l)$
\begin{equation}
  \yvect = \fvect(\xvect) = \left(\fvect^{(L - 1)}\circ\ldots\circ \fvect^{(0)} \right)(\xvect)  \,.
\end{equation}
Let $\Wvect$ be a collection of all model parameters (weights and biases) $W^{(l)}$ at all hidden layers. 
Each neuron computes its output as 
\begin{equation}
    f_i^{(l)} = \phi(\wvect_i^{(l)^\T}\fvect^{(l-1)})\,,
\end{equation}
where $\phi(\cdot)$ denotes a so-called activation function which introduces a nonlinearity at each layer.
Note that we absorbed the biases in $\wvect$.

Given a prior over $\Wvect$, the objective of Bayesian inference is to find the posterior distribution over all model parameters $\Wvect$ using the available input data $X = \{\xvect_1, \ldots, \xvect_n\}$ associated with labels $Y = \{\yvect_1, \ldots, \yvect_n\}$
\begin{equation}
    p(\Wvect|X, Y) = \frac{ p(Y|X,\Wvect) p(\Wvect) }{ p(Y|X) }\,.
\end{equation}
Bayesian inference for \dnns is analytically intractable and it is necessary to resort to approximations. 
One way to recover tractability is through the use of variational inference techniques as described next. 

\paragraph{Stochastic Variational Inference}
In variational inference, we introduce a family of distributions $q_{\thetavect}(\Wvect)$, parameterized through $\thetavect$, and attempt to find an element of this family which is as close to the posterior distribution of interest as possible \citep{Jordan99}.
This can be formulated as a minimization with respect to $\thetavect$ of the \KL divergence \citep{Kullback1959} between the elements of the family $q_{\thetavect}(\Wvect)$ and the posterior:
\begin{align}
    q_{\tilde{\thetavect}}(\Wvect) = \arg\min_{\thetavect} \{ \dkl{q_{\thetavect}(\Wvect)}{p(\Wvect | X, Y)} \} \,.
\end{align}
Simple manipulations allow us to rewrite this expression as the negative lower bound (\nelbo) to the marginal likelihood of the model (see supplementary material)
\begin{align} 
    \nelbo = \nll + \dkl{q_{\thetavect}(\Wvect)}{p(\Wvect)} \, , 
\end{align}
where the first term is the expected negative log-likelihood $\nll =\Exp_{q_{\thetavect}}\left[-\log p(Y|X, \Wvect)\right]$, and the second term acts as regularizer, penalizing distributions $q_{\thetavect}$ that deviate too much from the prior.
When the likelihood factorizes across data points, we can unbiasedly estimate the expectation term randomly selecting a mini-batch $\B$ of $m$ out of $n$ training points
\begin{equation}
    \nll \approx -\frac{n}{m}\sum_{\xvect, \yvect \in \B} \Exp_{q_{\thetavect}} \log p(\yvect|\xvect, \Wvect)\,.
\end{equation}
Each term in the sum can be further unbiasedly estimated using $N_{\mathrm{MC}}$ Monte Carlo samples as
\begin{equation}
    \Exp_{q_{\thetavect}} \log p(\yvect|\xvect, \Wvect) = \frac{1}{N_{\mathrm{MC}}}\sum_{i = 1}^{N_{\mathrm{MC}}} \log p(\yvect|\xvect, \Wvect_i)\,,
\end{equation}
where $\Wvect_i \sim q_{\thetavect}(\Wvect)$. 
Following \citet{Kingma14}, each sample $\Wvect_i$ is constructed using the reparameterization trick, which allows to obtain a deterministic dependence of the \nelbo w.r.t. $\thetavect$. 
Alternatively, it is possible to determine the distribution of the \dnn units $f_i^{(l)}$ before activation from $q_{\thetavect}(\Wvect)$. 
This trick, known as \textit{the local reparameterization trick}, allows one to considerably reduce the variance of the stochastic gradient w.r.t. $\thetavect$ and achieve faster convergence as shown by \citet{Kingma2015}.



\section{Proposed Method}
\label{sec:proposed-method}
\begin{figure*}[!t!]
    \centering
    \subfigure[]{
        \scalebox{0.75}{
        \ifdefined\compilefigures
        \tikzsetnextfilename{blm1}
        \input{../fig/methods/blm1}
        \else
        \includegraphics{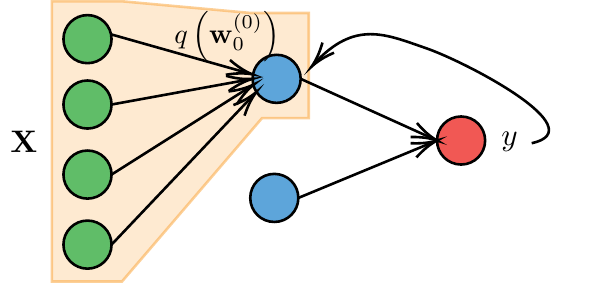}
        \fi
        }
    }
    \subfigure[]{
        \scalebox{0.75}{
        \ifdefined\compilefigures
        \tikzsetnextfilename{blm2}
        \input{../fig/methods/blm2}
        \else
        \includegraphics{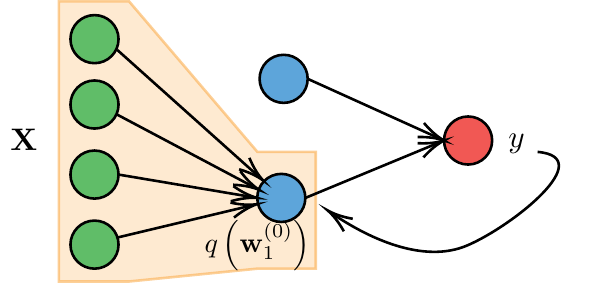}
        \fi
        }
    }
    \subfigure[]{
        \scalebox{0.75}{
        \ifdefined\compilefigures
        \tikzsetnextfilename{blm3}
        \input{../fig/methods/blm3}
        \else
        \includegraphics{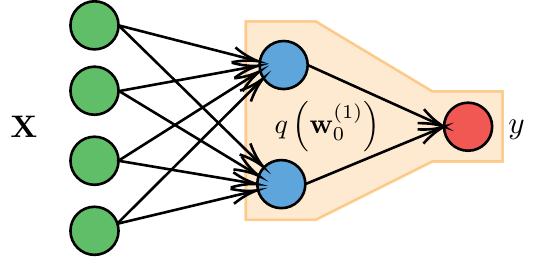}
        \fi
        }
    }
    
    \caption{Visual representation of the proposed method for initialization. In (a) and (b), we learn two Bayesian linear models, whose outputs are used in (c) to infer the following layer.}
    \label{fig:blm-demo}
\end{figure*}

In this section, we introduce our proposed Iterative Bayesian Linear Model (\iblm) initialization for \svi. 
We first introduce \iblm for regression with \dnns, and we then show how this can be extended to classification and to \cnns. 
\newcommand\mycommfont[1]{\tiny\ttfamily\textcolor{black}{#1}}
\SetCommentSty{mycommfont}
\SetKwComment{Comment}{$\triangleright$\ }{}
\begin{algorithm}[b!]
    
    \footnotesize
    \SetKwInOut{Input}{Inputs}
    \SetKwInOut{Output}{Returns}
    \Input{Model $M$, Dataset $D$}

    \ForEach{layer in $M$}{%
        \ForEach{outfeature in $layer$}{%
            $X, Y \gets$ next batch in $D$\;
            propagate $X$\;
            $X_{\texttt{BLM}} \gets$ output of previous layer\;
            \If(\Comment*[f]{ref \ref{subsec:iblm-cnn}}){$layer$ is convolutional}{%
                $X_{\texttt{BLM}} \gets$ patch extraction($X_{\texttt{BLM}}$)\;
            }
            \eIf(\Comment*[f]{ref \ref{subsec:iblm-class}}){$likelihood$ is classification}{
                $\text{var}(Y_{\texttt{BLM}}) \gets \log{[(Y + \alpha)^{-1} + 1]}$\;
                $\text{mean}(Y_{\texttt{BLM}}) \gets \log{(Y + \alpha) - \text{var}(Y_{\texttt{BLM}})/2}$\;
            }
            {
                $Y_{\texttt{BLM}} \gets Y$\;
            }
            $p(\wvect|X,Y) \gets$ BLM($X_{\texttt{BLM}}$,$Y_{\texttt{BLM}}$) \Comment*[r]{ref \ref{subsec:iblm-regr}}
            $q(\wvect) \gets$ best approx. of $p(\wvect|X,Y)$ \Comment*[r]{ref \ref{subsec:best_posterior}}
            
        }
    }
    
 \caption{Sketch of the \iblm Initializer}
    \label{alg:sketch-iblm}
\end{algorithm}

\subsection{Initialization of DNNs for Regression}
\label{subsec:iblm-regr}
In order to initialize the weights of \dnns, we proceed iteratively as follows. 
Before applying the nonlinearity through the activation function, each layer in a Bayesian \dnn can be seen as multivariate Bayesian linear regression model. 
We use this observation as an inspiration to initialize \svi as follows.
Starting from the first layer, we can set the parameters of $q(W^{(0)})$ by running Bayesian linear regression with inputs $X$ and labels $Y$. 
After this, we initialize the approximate posterior over the weights at the second layer $q(W^{(1)})$ by running Bayesian linear regression with inputs $X = \Phi(X \tilde{W}^{(0)})$ and labels $Y$. 
Here, $\Phi(\cdot)$ denotes the elementwise application of the activation function to the argument, whereas $\tilde{W}^{(0)}$ is a sample from $q(W^{(0)})$. 
We then proceed iteratively in the same way up to the last layer.
\autoref{fig:blm-demo} gives an illustration of the proposed method for a simple architecture. 

The intuition behind \iblm is as follows.
If one layer is enough to capture the complexity of a regression task, we expect to be able to learn an effective mapping right after the initialization of the first layer. 
In this case, we also expect that the mapping at the next layers implements simple transformations, close to the identity. 
Learning a set of weights with these characteristics starting from a random initialization is extremely hard; this motivated the work in \citet{He16} that proposed the residual network architecture. 
Our \iblm initialization takes this observation as an intuition to initialize \svi for general deep models.  

From a complexity point of view, denoting by $h^{(l)}$ the number of output neurons at layer $(l)$, this is equivalent to $h^{(l)}$ univariate Bayesian linear models. 
Instead of using the entire training set to learn the linear models, each one of these is inferred based on a random mini-batch of data, whose inputs are propagated through the previous layers.
The complexity of \iblm is linear in the batch size and cubic in the number of neurons to be initialized. 
Later on in this Section, we will provide an evaluation of the effect of batch size and a timing profiling of \iblm.




\subsection{From the Bayesian linear model posterior to the variational approximation} 
\label{subsec:best_posterior}
The proposed \iblm initialization of variational parameters can be used with any choice for the form of the approximate posterior.
The exact posterior of Bayesian linear regression is not factorized, so one needs to match this with the form of the chosen approximate posterior. 
For simplicity of notation, let $\wvect$ be the parameters of interest in Bayesian linear regression for a given output $\yvect = Y_{\cdot i}$. 
We can formulate this problem by minimizing the \KL divergence from $q(\wvect)$ to the actual posterior $p(\wvect | X, \yvect)$. 
In the case of a fully factorized approximate posterior over the weights $q(W^{(l)}) = \prod_{ij} q_{ij}(W^{(l)})$, this minimization can be done analytically.
This results in the mean being equal to the mean of $p(\wvect | X, \yvect)$ and the variances $(s_i^2)^{-1} = \Sigma^{-1}_{ii}$; see the supplementary material for the full derivation. 
Similar results can be also obtained for different posterior distributions, such as Gaussian posteriors with full or low-rank covariance, or matrix-variate Gaussian posteriors \citep{Louizos2016}.


\subsection{Initialization for Classification}
\label{subsec:iblm-class}
In this section we show how our proposal can be extended to $k$-class classification problems.  
We assume a one-hot encoding of the labels, so that $Y$ is an $n \times k$ matrix of zeros and ones (one for each row of $Y$).
Recently, it has been shown that it is possible to obtain an accurate modeling of the posterior over classification functions by applying regression on a transformation of the labels \citep{Milios2018}. 
This is interesting because it allows us to apply Bayesian linear regression as before in order to initialize \svi for \dnns. 

The transformation of the labels is based on the formalization of a simple intuition, which is the inversion of the softmax transformation. 
One-hot encoded labels are viewed as a set of parameters of a degenerate Dirichlet distribution. 
We resolve the degeneracy of the Dirichlet distribution by adding a small regularization, say $\alpha = 0.01$, to the parameters.
At this point, we leverage the fact that Dirichlet distributed random variables can be constructed as a ratio of Gamma random variables, that is, if $x_i \sim \mathrm{Gamma}(a_i, b)$, then $\frac{x_i}{\sum_j x_j}\sim \mathrm{Dir}(\avect)$.
%
We can then approximate the Gamma random variables with log-Normals by moment matching. 
By doing so, we obtain a representation of the labels which allows us to use standard regression with a Gaussian likelihood, and which retrieves an approximate Dirichlet when mapping predictions back using the softmax transformation. 
As a result, the latent functions obtained represent probabilities of class labels.


The only small complication is that the transformation imposes a different noise level for labels that are $0$ or $1$, and this is because of the non-symmetric nature of the transformation. 
Nevertheless, it is a simple matter to extend Bayesian linear regression to handle heteroscedasticity; see the supplement for a derivation of heteroscedastic Bayesian linear regression and \citet{Milios2018} for more insights on the transformation to apply regression on classification problems.

\subsection{Initialization of CNNs}
\label{subsec:iblm-cnn}
The same method can be also applied on \cnns. 
Convolutional layers are commonly implemented as matrix multiplication (e.g. as a linear model) between a batched matrix of patches and a reshaped filter matrix \citep{Jia2014}. 
Rather than using the outputs of the previous layer as they are, for convolutional layers each Bayesian linear model learns the mapping from spatial patches to output features. 
Finally, in \autoref{alg:sketch-iblm} we summarize a sketch of the proposed method for regression as well as for classification and convolutional layers. 


\subsection{General Insights on \iblm}

\begin{figure}[!t]
    \centering
    \setlength\figureheight{.135\textwidth}
    \setlength\figurewidth{.35\textwidth} 
    \hspace{-5mm}
    \subfigure{
        \tiny
        \ifdefined\compilefigures
        \tikzsetnextfilename{bound_iblm}
        \input{../fig/batchsize-exp/bound_iblm2}
        \else
        \includegraphics{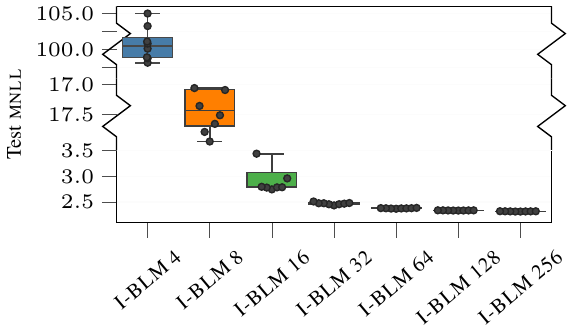}
        \fi
    \label{fig:bound_iblm}
    }
    \setlength\figureheight{.22\textwidth}
    \setlength\figurewidth{.18\textwidth} 
    \hspace{-6mm}
    \subfigure{
        \tiny
        \ifdefined\compilefigures
        \tikzsetnextfilename{bound_iblm_vs_mcd}
        \input{../fig/batchsize-exp/bound_iblm_vs_mcd}
        \else
        \includegraphics{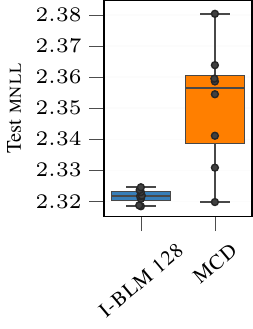}
        \fi
    \label{fig:bound_iblm_vs_mcd}
    }
    \caption{Comparison of test \mnll after initialization of \lenet for \mnist averaged out of eight successive runs. On the \textbf{left}, with different batch sizes, on the \textbf{right} with \mcd.}
\end{figure}
\begin{figure}[!b]%
    \centering
    \setlength\figureheight{.23\textwidth}
    \setlength\figurewidth{.27\textwidth} 
    \subfigure{
        \tiny
        \ifdefined\compilefigures
        \tikzsetnextfilename{time_init_lenet}
        \input{../fig/batchsize-exp/time_init_lenet}
        \else
        \includegraphics{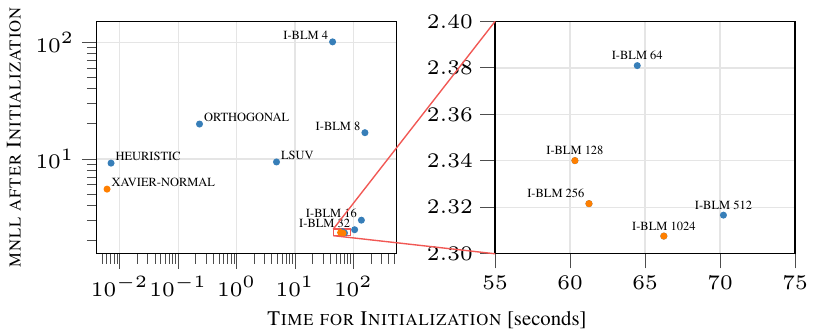}
        \fi
    }
    \caption{On the \textbf{left}, comparison of initialization time versus test \mnll, averaged out of eight successive runs (on the \textbf{right},  magnification of the small portion of the plot). Orange corresponds to the Pareto frontier.}
    \label{fig:time_init_lenet}
\end{figure}

Previously we claimed that (i) small batches of data are sufficient to solve the Bayesian linear model and that (ii) our initialization does not suffer from timing complexity overhead.   
We now want to justify the aforementioned claims.
We initialize a \cnn (\lenet) on \mnist with an increasing number of samples per batch; Figure \autoref{fig:bound_iblm} shows how test log-likelihood is affected by this choice.
Using the full training set leads to a better estimate of the posteriors and therefore of the likelihood. 
The mini-batch size affects also the heterogeneity of the posteriors, which vanishes when using the full training set.
Nonetheless, we show that from 64/128 samples the improvement on the test \mnll is only marginal.
The same experiment is also repeated comparing test \mnll after initialization between \svi with \iblm and \mcd (Figure \autoref{fig:bound_iblm_vs_mcd}).
Similar comments apply also for this case: \iblm allows the training to start from a lower negative log-likelihood.
Finally, \autoref{fig:time_init_lenet} reports the test \mnll after initialization as a function of the time required (orange points correspond to Pareto-optimal points). Before training, three out of four optimal initializers are \iblm.

\section{Experimental Results}
\label{sec:results}

In this section, we compare different initialization algorithms for \svi to prove the effectiveness of \iblm.
We propose a number of competitors inspired from the literature developed for loss minimization in \dnns and \cnns. 
In the case of \cnns, we also compare with Monte Carlo Dropout (\mcd; \citet{Gal16}) and Natural Noisy Gradients (\textsc{noisy-kfac}; \citet{Zhang2018}), which respresent the state-of-the-art references for inference in Bayesian \cnns.
At layer $(l)$, given the fully-factorized variational distribution $q(W^{(l)}) = \prod_{i, j}\N(w_{i, j}^{(l)}\vert \mu_{i, j}, \sigma^2_{i, j})$, we initialize $\mu_{i, j}$ and $\sigma^2_{i, j}$ with the following methods.

\paragraph{Uninformative} The optimization of the posterior starts from the prior; therefore $q(W) = \prod_{i, j}\N(w_{i, j}\vert 0, 1)$. Note that this yields an initial \KL divergence in the \nelbo equal to zero. 

\paragraph{Random Heuristic} An extension to commonly used heuristic with $\mu_{i, j} = 0$ and $\sigma^2_{i, j} = \frac{1}{\Din}$, with $\Din$ the number of input features at layer $(l)$.

\paragraph{Xavier Normal} Originally proposed by \citet{Glorot2010}, it samples all weights independently from a Gaussian distribution with zero mean and $\sigma^2 = \frac{2}{\Din + \Dout}$. 
This variance-based scaling avoids issues with vanishing or exploding gradients. 
In this case, it is straightforward to extend it to the case of \svi; indeed, instead of sampling, we directly set $\mu_{i, j} = 0$ and $\sigma^2_{i, j} = \frac{2}{\Din + \Dout}$, knowing that the sampling is performed during the Monte Carlo estimate of the log-likelihood.

\paragraph{Orthogonal} Starting from an analysis of learning dynamics of \dnns with linear activations, \citet{Saxe2013} propose an initialization scheme with orthonormal weight matrices. 
The idea is to decompose a Gaussian random matrix onto an orthonormal basis, and use the resulting orthogonal matrix. 
We adapt this method for \svi by initializing the mean matrix with the orthogonal matrix and $\sigma^2_{i, j} = \frac{1}{\Din}$. 
For our experiments, we use the implementation in \pytorch \citep{Paszke2017} provided by the Authors, which uses a QR-decomposition.

\begin{figure}[!b!]
    \centering
    \setlength\figureheight{.18\textwidth}
    \setlength\figurewidth{.27\textwidth} 
    \hspace{-0.01\textwidth}
    \subfigure{
        \tiny
        \ifdefined\compilefigures
        \tikzsetnextfilename{regression-shallow-powerplant}
        \input{../fig/regr-exp-gpu/powerplant.tex}
        \else
        \includegraphics{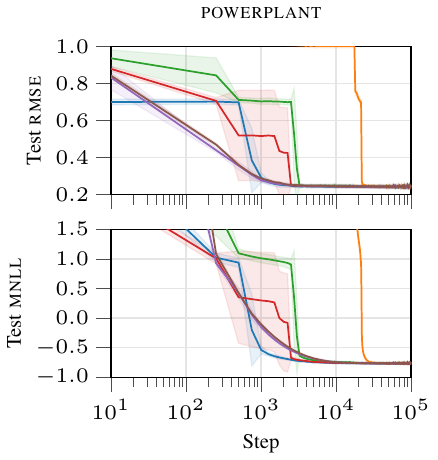}
        \fi
    }
    \hspace{-0.03\textwidth}
    \subfigure{
        \tiny
        \ifdefined\compilefigures
        \tikzsetnextfilename{regression-shallow-protein}
        \input{../fig/regr-exp-gpu/protein.tex}
        \else
        \includegraphics{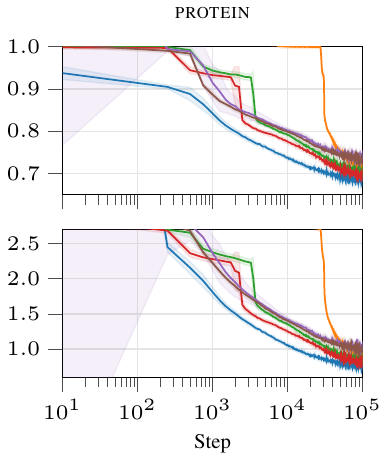}
        \fi
    }
    \\[-1ex]
    \subfigure{
        \tiny
        \ifdefined\compilefigures
        \tikzsetnextfilename{regression-shallow-legend}
        \input{../fig/regr-exp-gpu/legend}
        \else
        \includegraphics{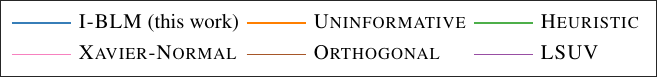}
        \fi
    }
    \caption{Progression of test \rmse and test \mnll  with different initializations on a shallow architecture.}
    \label{fig:regression-shallow}
\end{figure}

\paragraph{Layer-Sequential Unit-Variance (LSUV)} Starting from orthogonal initialization, \citet{Mishkin2015} propose to perform a layer sequential variance scaling of the weight matrix. 
By implementing a data-driven greedy initialization, it generalizes the results to any nonlinear activation function and even to any type of layers that can impact the variance of the activations. 
We implement Layer-Sequential Unit-Variance (\lsuv) for the means, while the variances are set to $\sigma^2_{i, j} = \frac{1}{\Din}$.

\subsection{Experiments}


Throughout the experiments, we use \adam optimizer \citep{Kingma2015b} with learning rate $10^{-3}$, batch size $64$, and $16$ Monte Carlo samples at training time and $128$ at test time.
All experiments are run on a server equipped with two 16c/32t Intel Xeon CPU and four NVIDIA Tesla P100, with a maximum time budget of 24 hours (never reached). 
To better understand the effectiveness of different initialization, all learning curves are plotted w.r.t. training iteration rather than wall-clock time.


\paragraph{Regression with a shallow architecture}

In this experiment we compare initialization methods for a shallow \dnn architecture on two datasets. 
The architecture used in these experiments has one single hidden layer with $100$ hidden neurons and \relu activations. 
We impose that the approximate posterior has fully factorized covariance. 
\autoref{fig:regression-shallow} shows the learning curves on the \powerplant ($n=9568$, $d=4$) and \protein ($n=45730$, $d=9$) datasets, repeated over five different train/test splits. 
\iblm allows for a better initialization compared to the competitors, leading to a lower root mean square error (\rmse) and lower mean negative log-likelihood (\mnll) on the test for a given computational budget. 
We refer the reader to the supplementary material for a more detailed analysis of the results. 

\paragraph{Regression with a deeper architecture}
\begin{figure}[!b!]
    \centering
    \setlength\figureheight{.18\textwidth}
    \setlength\figurewidth{.27\textwidth} 
    \hspace{-0.015\textwidth}
    \subfigure{
        \tiny
        \ifdefined\compilefigures
        \tikzsetnextfilename{regression-deep-powerplant}
        \input{../fig/regr-depth-exp/powerplant.tex}
        \else
        \includegraphics{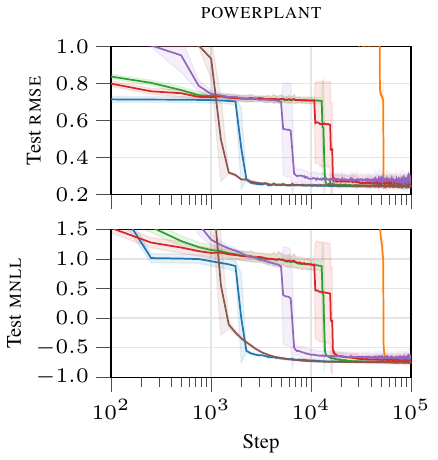}
        \fi
    }
    \hspace{-0.025\textwidth}
    \subfigure{
        \tiny
        \ifdefined\compilefigures
        \tikzsetnextfilename{regression-deep-protein}
        \input{../fig/regr-depth-exp/protein.tex}
        \else
        \includegraphics{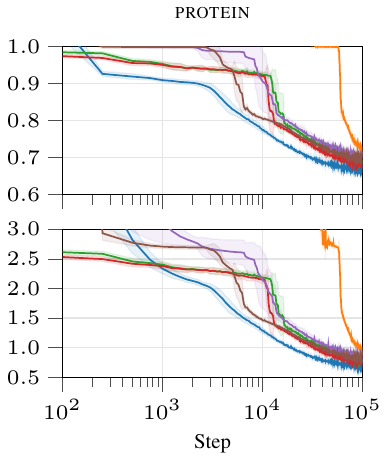}
        \fi
    }
    \\[-1ex]
    \subfigure{
        \tiny
        \ifdefined\compilefigures
        \tikzsetnextfilename{regression-deep-legend}
        \input{../fig/regr-depth-exp/legend}
        \else
        \includegraphics{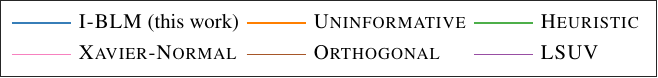}
        \fi
    }
    \caption{Progression of test \rmse and test \mnll with different initializations on a deep architecture.}
    \label{fig:regression-deep}
\end{figure}

Similar considerations hold when increasing the depth of the model, keeping the same experimental setup. 
\autoref{fig:regression-deep} shows the progression of the \rmse and \mnll error metrics when using \svi to infer parameters of a \dnn with five hidden layers and $100$ hidden neurons per layer, and \relu activations.
Again, the proposed initialization allows \svi to converge faster than when using other initializations. 

\paragraph{Classification with a deep architecture}

\begin{figure*}[!t]
    \centering
    \setlength\figureheight{.18\textwidth}
    \setlength\figurewidth{.28\textwidth} 
    \hspace{-0.045\textwidth}
    \subfigure{
        \tiny
        \ifdefined\compilefigures
        \tikzsetnextfilename{classification-deep-mnist}
        \input{../fig/class-depth-exp/mnist.tex}
        \else
        \includegraphics{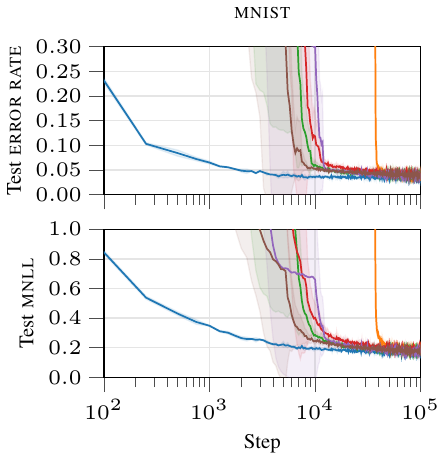}
        \fi
    }
    \hspace{-0.025\textwidth}
    \subfigure{
        \tiny
        \ifdefined\compilefigures
        \tikzsetnextfilename{classification-deep-eeg}
        \input{../fig/class-depth-exp/eeg.tex}
        \else
        \includegraphics{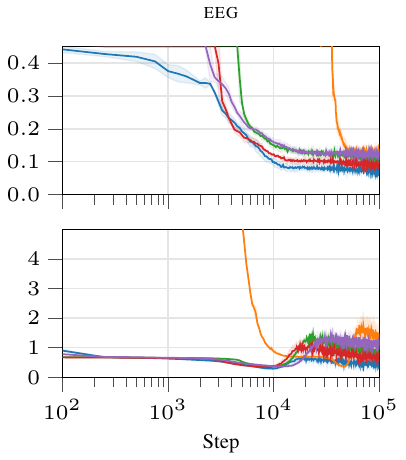}
        \fi
    }
    \hspace{-0.025\textwidth}
    \subfigure{
        \tiny
        \ifdefined\compilefigures
        \tikzsetnextfilename{classification-deep-credit}
        \input{../fig/class-depth-exp/credit.tex}
        \else
        \includegraphics{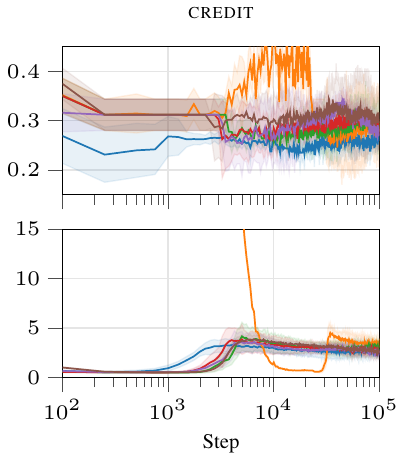}
        \fi
    }
    \hspace{-0.025\textwidth}
    \subfigure{
        \tiny
        \ifdefined\compilefigures
        \tikzsetnextfilename{classification-deep-spam}
        \input{../fig/class-depth-exp/spam.tex}
        \else
        \includegraphics{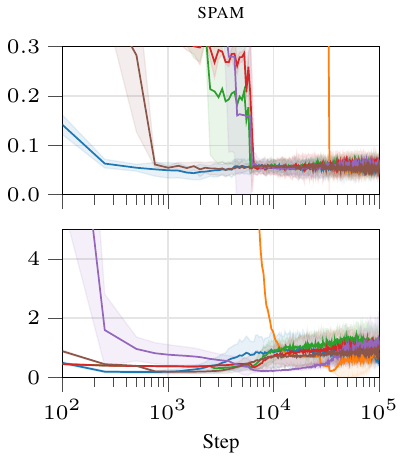}
        \fi
    }
    \\[-1ex]
    \subfigure{
        \tiny
        \ifdefined\compilefigures
        \tikzsetnextfilename{classification-deep-legend}
        \input{../fig/class-depth-exp/legend}
        \else
        \includegraphics{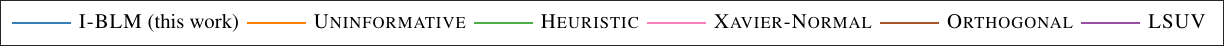}
        \fi
    }
    \caption{Progression of test \errate and test \mnll with different initializations on classification problems.}
\end{figure*}

Using the same deep \dnn architecture as in the last experiment (five hidden layers with $100$ neurons), we tested \iblm with classification problems on \mnist ($n=70000$, $d=784$), \eeg ($n=14980$, $d=14$), \credit ($n=1000$, $d=24$) and \spam ($n=4601$, $d=57$). 
Interestingly, with this architecture, some initialization strategies struggled to converge, e.g., \uninformative on \mnist and \lsuv on \eeg. 
The gains offered by \iblm achieves the most striking results on \mnist. 
After less than $1000$ training steps (less than an epoch), it can already reach a test accuracy greater than $95\%$; other initalizations reach such performance much later during training.
Even after $100$ epochs, \svi inference initialized with \iblm provides on avarage an increase up to $14\%$ of accuracy at test time. 
Full results are reported in the supplementary material.

\paragraph{Experiments on CNNs}
\begin{figure}[!t!]
    \centering
    \setlength\figureheight{.18\textwidth}
    \setlength\figurewidth{.27\textwidth} 
    \hspace{-0.01\textwidth}
    \subfigure{
        \tiny
        \ifdefined\compilefigures
        \tikzsetnextfilename{convnet-mnist}
        \input{../fig/convnet-exp/mnist}
        \else
        \includegraphics{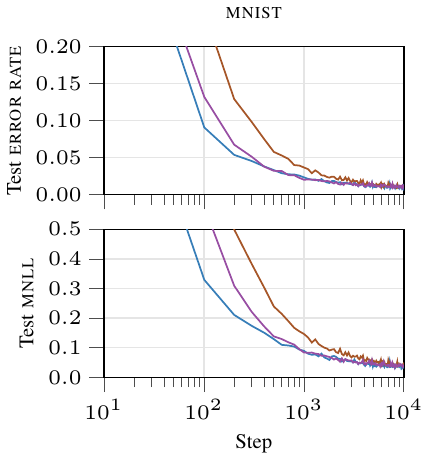}
        \fi
    }
    \hspace{-0.025\textwidth}
    \subfigure{
        \tiny
        \ifdefined\compilefigures
        \tikzsetnextfilename{convnet-cifar10}
        \input{../fig/convnet-exp/cifar10}
        \else
        \includegraphics{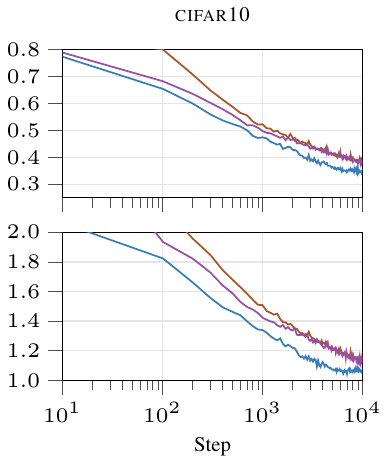}
        \fi
    }
    \\[-1ex]
    \subfigure{
        \tiny
        \ifdefined\compilefigures
        \tikzsetnextfilename{convnet-legend}
        \input{../fig/convnet-exp/legend}
        \else
        \includegraphics{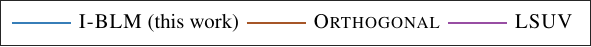}
        \fi
    }
    \caption{Progression of test \errate and test \mnll for different initializations using \lenet on \mnist and \cifar. NB: \uninformative, \heuristic and \xavier could never converged and therefore they are removed.}
    \label{fig:convnet}
\end{figure}

For this experiment, we implemented a Bayesian version of the original \lenet architecture proposed by \citet{LeCun1998} with two convolutional layers of $6$ and $16$ filters, respectively and \relu activations applied after all convolutional layers and fully-connected layers.
We tested our framework on \mnist and on \cifart.
The only initialization strategies that achieve convergence are \orthogonal and \lsuv, along with \iblm; the other methods did not converge, meaning that they push the posterior back to the prior.
\autoref{fig:convnet} reports the progression of \errate and \mnll.
For both \mnist and \cifart, \iblm places the parameters where the network can consistently deliver better performance both in terms of \errate and \mnll throughout the entire learning procedure.


\paragraph{Comparison with large scale models and non-Gaussian approximation}

\begin{figure}[!t!]
    \centering
    \setlength\figureheight{.18\textwidth}
    \setlength\figurewidth{.27\textwidth} 
    \hspace{-0.01\textwidth}
    \subfigure{
        \tiny
        \ifdefined\compilefigures
        \tikzsetnextfilename{lenet_mcd_mnist}
        \input{../fig/montecarlo-exp/mnist}
        \else
        \includegraphics{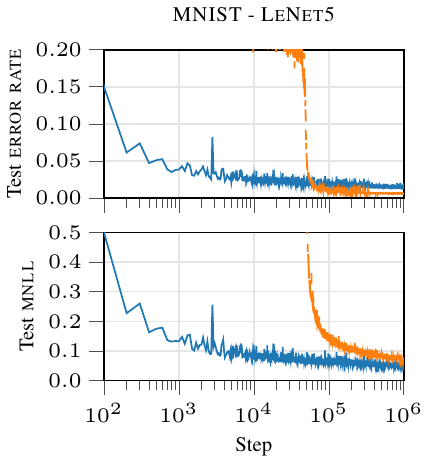}
        \fi
    }
    \hspace{-0.025\textwidth}
    \subfigure{
        \tiny
        \ifdefined\compilefigures
        \tikzsetnextfilename{alexnet-curves}
        \input{../fig/alexnet-exp/alexnet}
        \else
        \includegraphics{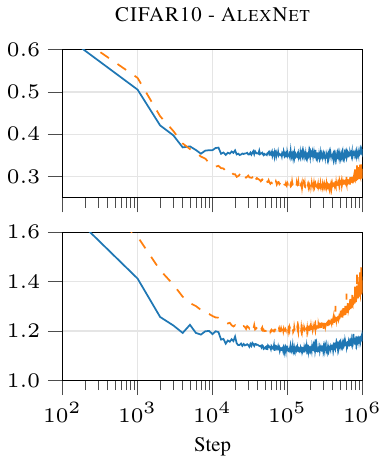}
        \fi
    }
    \\[-1ex]
    {
        \tiny
        \ifdefined\compilefigures
        \tikzsetnextfilename{lenet_mcd_legend}
        \input{../fig/montecarlo-exp/legend}
        \else
        \includegraphics{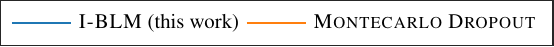}
        \fi
    }
    \caption{Comparison between \svi with Gaussian  approximation and \mcd on \mnist and \cifart with \lenet and \alexnet.}
    \label{fig:mcd-curves}

\end{figure}

Monte Carlo Dropout (\mcd; \citet{Gal16b}) offers a simple and effective way to perform approximate Bayesian \cnn inference, thanks to the connection that the Authors have established between dropout and variational inference. 
In this experiment, we aim to compare and discuss benefits and disadvantages of using a Gaussian posterior approximation with respect to the Bernoulli approximation that characterizes \mcd. 
For a fair comparison, we implemented the same \lenet architecture and the same learning procedure in \citet{Gal16b}\footnote{\url{https://github.com/yaringal/DropoutUncertaintyCaffeModels}}. 
In particular, for \mnist, the two convolutional layers have $20$ and $50$ filters, respectively. 
Dropout layers are placed after every convolutional and fully-connected layers with a dropout probability of $0.5$. 
To replicate the results in \citet{Gal16b}, we used the same learning rate policy $\texttt{base-lr} \times (1 + \xi \times \texttt{iter})^{-p}$ with $\xi = 0.0001$, $p = 0.75$, $\texttt{base-lr}=0.01$ and weight decay of $0.0005$.
\autoref{fig:mcd-curves} shows the learning curves. 
Monte Carlo Dropout achieves state-of-art \errate but the form assumed by \mcd for the posterior is reflected on an higher \mnll compared to \svi with a Gaussian posterior. 
Provided with a nontrivial initialization, Gaussian \svi can better fit the model and deliver a better quantification of uncertainty. 
%

We report also \errate and \mnll for \svi with \iblm and \mcd on \alexnet \citep{Krizhevsky2012}. 
The \cnn is composed by a stack of five convolutional layers and three fully-connected layers for a total of more than $1$M parameters ($2$M for \svi).
In this experiment, we have experienced the situation in which, due to the overparameterization of the model, the \nelbo is completely dominated by the \KL divergence. 
Therefore, the prior has a large influence on the optimization, so we decided to follow the approach in \citet{Graves11}, allowing for a phase of optimization of the variances of the prior over the parameters.
The results are reported in \autoref{fig:mcd-curves}. 
Once again, we show that \svi with \iblm provides a lower negative log-likelihood with respect to Bernoulli approximation in \mcd. 


Finally, we demonstrate that -- provided with a sensible initialization -- even simple factorized Gaussian posterior can achieve state-of-the-art performance on \cifart with \vgg\footnotemark, a large scale \cnn  \citep{Simonyan14}.
\footnotetext{The architecture implemented is the same as in \citet{Zhang2018}: 32-32-M-64-64-M-128-128-128-M-256-256-256-M-256-256-256-M-FC10}
In this experiment, in addition to \mcd, we compare with also with \textsc{noisy-kfac}, an approximation of matrix-variate Gaussian posterior using noisy natural gradients introduced by \citet{Zhang2018}. 
In the case of Gaussian \svi, to deal with over-parameterization, we implemented a policy where the \KL term is gradually included in the \nelbo (more details in the supplementary material).
Results are shown in \autoref{fig:vgg-plots} and in the adjacent Table.
Gaussian \svi delivers state-of-art test \mnll while also providing a competitive test \errate.

\begin{figure}[!t]
    \centering
  \begin{minipage}[c]{0.265\textwidth}
    \centering
    \setlength\figureheight{.7\textwidth}
    \setlength\figurewidth{1.1\textwidth} 
    \hspace{-0.01\textwidth}
        \tiny
    \ifdefined\compilefigures
        \tikzsetnextfilename{vgg16_iblm_vs_mcd_vs_noisykfac}
        \input{../fig/vgg-exp/vgg16_iblm_vs_mcd_vs_noisykfac}
    \else
        \includegraphics{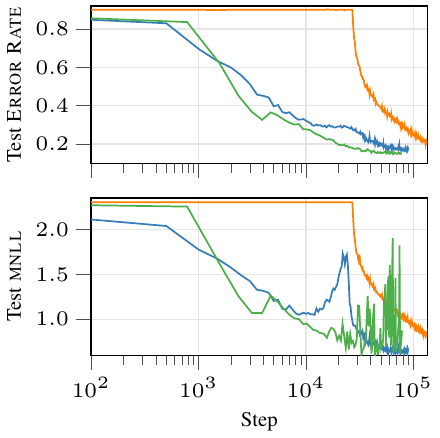}
    \fi
    \\
  \end{minipage}
  \begin{minipage}[c]{0.21\textwidth}
    \centering
    \tiny
    \begin{tabular}[t]{r||c|c}
        \tiny 
         & \mnll & \textsc{error} \\
         \midrule
        \iblm   & $\mathbf{0.6443}$ & $\mathbf{0.1682}$\\
        \mcd          & $0.8213$ & $0.2147$ \\
      \textsc{n-kfac} & $0.7327$\footnote{\scriptsize Due to high variance in the test \mnll, in this case we report the average of the last 5 epochs.} & $\mathbf{0.1506}$ \\
    \end{tabular}
    \captionlistentry[table]{A table beside a figure}
    \end{minipage}
    \vspace{0.1mm}
    \ifdefined\compilefigures
    \tikzsetnextfilename{vgg_legend}
    \input{../fig/vgg-exp/legend}
    \else
    \includegraphics{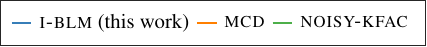}
    \fi
    \captionsetup{labelformat=andtable}
    \captionof{figure}{Comparison between Gaussian factorized \svi, \mcd and \textsc{noisy-kfac} on \vgg with \cifart}
    \label{fig:vgg-plots}
\end{figure}

\paragraph{Extended experimental evaluation}
We refer the reader to the supplementary material for additional insights of \iblm compared with other initialization methods, analysis of out-of-sample uncertainty estimation, and tests of calibration properties of deep classifiers. 





%
%
%
%


\section{Conclusions}
\label{sec:conclusions}


This work fills an important gap in the literature of Bayesian deep learning, that is how to effectively initialize variational parameters in \svi. 
We proposed a novel way to do so, \iblm, which is based on an iterative layer-wise initialization based on Bayesian linear models. 
Through a series of experiments, including regression and classification with \dnns and \cnns, we demonstrated the ability of our approach to consistently initialize the optimization in a way that makes convergence faster than alternatives inspired from the state-of-the-art in loss minimization for deep learning. 

Thanks to \iblm, it was possible to carry out an effective comparison with state-of-the-art methods to carry out approximate inference for \dnns and \cnns. 
This suggests a number of directions to investigate to improve on \svi and Bayesian \cnns. 
We found that the choice of the prior plays an important role in the behavior of the optimization, so we are investigating ways to define sensible priors for these models.
Furthermore, we are looking into ways to extend our initialization strategy to more complex posterior distributions beyond Gaussian and to other deep models, such as Deep Gaussian Processes and Bayesian deep generative models.

\paragraph{Acknowledgements}
    MF gratefully acknowledges support from the AXA Research Fund.


\FloatBarrier
\clearpage



 \appendix
 \onecolumn

 \section{Full derivation of variational lower bound}
 \label{appex:dkl-derivation}
\newcommand{\highlight}[1]{%
\colorbox{white!50}{$\displaystyle#1$}}

\begin{align*} 
  \dkl{q_{\theta}(\Wvect)}{p(\Wvect|X,Y)} &=
  \Exp_{q_{\theta}} \left[ \log\frac{q_\theta(\Wvect)}{p(\Wvect|X, Y)} \right] = \\
  &= \Exp_{q_{\theta}}\left[\log q_{\theta}(\Wvect) - \log p(\Wvect|X, Y)\right] =  \\
  &= \Exp_{q_{\theta}}\left[-\log p(Y|X, \Wvect)\right] +\Exp_{q_{\theta}}\left[\log q_{\theta}(\Wvect) - \log p(\Wvect) \right] + \log p(Y|X) = \\
  &= \highlight{\nll +  \dkl{q_{\theta}(\Wvect)}{p(\Wvect)}} + \log p(Y|X) 
\end{align*}

 \section{Bayesian linear regression}
 We express the likelihood and the prior on the parameters as follows:
$$
p(Y | W, L) = \prod_{i} p(Y_{\cdot i} | X W_{\cdot i}, L) = \prod_{i} \norm(Y_{\cdot i} | X W_{\cdot i}, L)
$$
Denote by $X$ the $n \times d$ matrix containing $n$ input vectors $\mathbf{x}_i \in \mathbb{R}^d$, and let $Y$ be the set consisting of the corresponding multivariate labels $\yvect_i$. 
In Bayesian linear regression we introduce a set of latent variables that we compute as a linear combination of the input through a set of weights, and we express the likelihood and the prior on the parameters as follows:
$$
p(Y | W, L) = \prod_{i} p(Y_{\cdot i} | X W_{\cdot i}, \lambda) = \prod_{i} \norm(Y_{\cdot i} | X W_{\cdot i}, L)
$$
and
$$
p(W | \Lambda) = \prod_{i} p(W_{\cdot i})  = \norm(W_{\cdot i} | \zerovect, \Lambda)
$$
The posterior of this model is:
$$
p(W | Y, L) \propto \prod_{i} \norm(Y_{\cdot i} | X W_{\cdot i}, L) \norm(W_{\cdot i} | \zerovect, \Lambda)
$$
which implies that the posterior factorizes across the columns of $W$, with factors
$$
p(W_{\cdot i} | Y, X, L, \Lambda) = 
\norm(W_{\cdot i} | \Sigma_i X^{\top} L^{-1} Y_{\cdot i}, \Sigma_i) 
$$
with $\Sigma_i = (\Lambda^{-1} + X^{\top} L^{-1} X)^{-1}$.
Similarly, the marginal likelihood factorizes as the product of the following factors
$$
p(Y_{\cdot i} | X, L, \Lambda) = 
\norm(Y_{\cdot i} | \zerovect, L + X \Lambda X^{\top}) 
$$

 \section{Heteroscedastic Bayesian linear regression}
 
We can extend Bayesian linear regression to the heteroscedastic case where $L = \diag(\sigmavect^2)$ and $\Lambda = I$.
These yield
$$
p(W_{\cdot i} | Y, X, \sigmavect^2) = 
\norm(W_{\cdot i} | \mu_i, \Sigma_i) 
\quad
\mathrm{with}
$$
$$
\mu_i = \Sigma_i X^{\top} \diag(\sigmavect^{-2}) Y_{\cdot i}
$$
$$
\Sigma_i = (I + X^{\top} \diag(\sigmavect^{-2}) X)^{-1}
$$
and
$$
p(Y_{\cdot i} | X, \sigmavect^2) = 
\norm(Y_{\cdot i} | \zerovect, \diag(\sigmavect^{2}) + X X^{\top}) 
$$

The expression for the marginal likelihood is computationally unconvenient due to the need to deal with an $n \times n$ matrix. 
We can use Woodbury identities\footnote{
$| I + B^{\top} C | = | I + C B^{\top} |$
and 
$
(A + UCV)^{-1} = A^{-1} - A^{-1} U ( C^{-1} + V A^{-1} U)^{-1} V A^{-1}
$
}
to express this calculation using $\Sigma_i$. 
In particular, 
$$
\log[p(Y_{\cdot i} | X, \sigmavect^2)] = -\frac{1}{2} \log \left\vert \diag(\sigmavect^{2}) + X X^{\top} \right\vert - \frac{1}{2} Y_{\cdot i}^{\top} \left(\diag(\sigmavect^{2}) + X X^{\top}\right)^{-1} Y_{\cdot i} + \mathrm{const.}
$$

Using Woodbury identites, we can rewrite the algebraic operations as follows:
$$
\log \left\vert \diag(\sigmavect^{2}) + X X^{\top} \right\vert = 
\log \left\vert \diag(\sigmavect^{2}) \right\vert + \log \left\vert I + \diag(\sigmavect^{-2}) X X^{\top} \right\vert =
\sum_j \log \sigma^2_j + \log \left\vert I +  X^{\top} \diag(\sigmavect^{-2}) X \right\vert
$$
and
$$
\left(\diag(\sigmavect^{2}) + X X^{\top}\right)^{-1} = 
\diag(\sigmavect^{-2}) - \diag(\sigmavect^{-2}) X \left(I +  X^{\top} \diag(\sigmavect^{-2}) X\right)^{-1} X^{\top} \diag(\sigmavect^{-2})
$$

So, wrapping up, we ca express all quantities of interest as:
$$
\Sigma_i^{-1} = I + X^{\top} \diag(\sigmavect^{-2}) X
$$
$$
\log[p(Y_{\cdot i} | X, \sigmavect^2)] = -\frac{1}{2} 
\left( 
\sum_j \log \sigma^2_j + \log\left\vert \Sigma_i^{-1} \right\vert
\right)
- \frac{1}{2} Y_{\cdot i}^{\top} \left(
\diag(\sigmavect^{-2}) - \diag(\sigmavect^{-2}) X \Sigma_i X^{\top} \diag(\sigmavect^{-2})
\right)^{-1} Y_{\cdot i} + \mathrm{const.}
$$
If we factorize $\Sigma_i^{-1} = Q Q^{\top}$, we obtain:
$$
\log[p(Y_{\cdot i} | X, \sigmavect^2)] = -\frac{1}{2} 
\left( 
\sum_j \log(\sigma^2_j) + \sum_k 2 \log(Q_{kk})
\right)
- \frac{1}{2} Y_{\cdot i}^{\top} \tilde{Y}_{\cdot i}
+ \frac{1}{2} 
\tilde{Y}_{\cdot i}^{\top} X Q^{-\top} Q^{-1} X^{\top} \tilde{Y}_{\cdot i} + \mathrm{const.}
$$
where $\tilde{Y}_{\cdot i} = \diag(\sigmavect^{-2}) Y_{\cdot i}$

Predictions follow from the same identities as before - looking at the predicted latent process, we have
$$
p(f_{*i} | X, Y, \xvect_{*}) = \int p(f_{*i} | W, \xvect_{*}) p(W | X, Y) dW
$$
We can again remove the dependence from the dimensions of $W$ that do not affect the prediction for the $i$th function as
$$
p(f_{*i} | X, Y, \xvect_{*}) = \int p(f_{*i} | W_{\cdot i}, \xvect_{*}) p(W_{\cdot i} | X, Y) dW
$$
Now:
$$
p(f_{*i} | W_{\cdot i}, \xvect_{*}) = \norm(f_{*i} | \xvect_*^{\top} W_{\cdot i}, 0)
\qquad \text{and} \qquad
p(W_{\cdot i} | X, Y) = \norm(W_{\cdot i} | \mu_i, \Sigma_i)
$$
giving
$$
p(f_{*i} | X, Y, \xvect_{*}) = \norm(f_{*i} | \xvect_*^{\top} \mu_i, \xvect_*^{\top} \Sigma_i \xvect_*)
$$

 \section{Full derivation of fully factorized Gaussian posterior approximation to Bayesian linear regression posterior}
 For simplicity of notation, let $\wvect$ be the parameters of interest in Bayesian linear regression for a given output $\yvect = Y_{\cdot i}$. 
We can formulate the problem of obtaining the best approximate factorized posterior of a Bayeian linear model as a minimization of the Kullback-Leibler divergence between $q(\wvect) = \norm(\wvect | \mvect, \diag(\svect^2))$ and the actual posterior $p(\wvect | X, \yvect)$. 
The expression of the KL divergence between multivariate Gaussians $p_0 = \norm(W | \muvect_0, \Sigma_0)$ and $p_1 = \norm(W | \muvect_1, \Sigma_1)$ is as follows:
$$
\mathrm{KL}[p_0 || p_1] = 
\frac{1}{2} \Tr(\Sigma_1^{-1} \Sigma_0)
+ \frac{1}{2} (\muvect_1 - \muvect_0)^{\top} \Sigma_1^{-1} (\muvect_1 - \muvect_0)
- \frac{D}{2} 
+ \frac{1}{2} \log\left( \frac{\mathrm{det}\Sigma_1}{\mathrm{det}\Sigma_0} \right)
$$

The KL divergence is not symmetric, so the order in which we take this matters. 
In case we consider $\mathrm{KL}[p(\wvect | X, \yvect) || q(\wvect)]$, the expression becomes:
$$
\mathrm{KL}[p(\wvect | X, \yvect) || q(\wvect)] = 
\frac{1}{2} \Tr(\diag(\svect^2)^{-1} \Sigma)
+ \frac{1}{2} (\mvect - \muvect)^{\top} \diag(\svect^2)^{-1} (\mvect - \muvect)
- \frac{D}{2} 
+ \frac{1}{2} \log\left( \frac{\prod_i s_i^2}{\mathrm{det}\Sigma} \right)
$$
It is a simple matter to show that the optimal mean $\mvect$ is $\muvect$ as $\mvect$ appears only in the quadratic form which is clearly minimized when $\mvect = \muvect$. 
For the variances $\svect^2$, we need to take the derivative of the KL divergence and set it to zero:
$$
\frac{\partial \mathrm{KL}[p(\wvect | X, \yvect) || q(\wvect)]}{\partial s^2_i} = 
 \frac{1}{2} \frac{\partial \Tr(\diag(\svect^2)^{-1} \Sigma)}{\partial s_i^2}
+ \frac{1}{2} \frac{\partial \sum_i \log{s_i^2}}{\partial s_i^2}
= 0
$$
Rewriting the trace term as the sum of the Hadamrd product of the matrices in the product $\sum_{ij} (\diag(\svect^2)^{-1} \odot \Sigma)_{ij} = \sum_i \Sigma_{ii} / s_i^2$, this yields 
$$
\frac{\partial \mathrm{KL}[p(\wvect | X, \yvect) || q(\wvect)]}{\partial s^2_i} = 
 \frac{1}{2} \frac{\partial \Sigma_{ii} / s_i^2 )}{\partial s_i^2}
+ \frac{1}{2} \frac{\partial \log{s_i^2}}{\partial s_i^2}
= 0
$$
This results in $s_i^2 = \Sigma_{ii}$, which is the simplest way to approximate the correlated posterior over $\wvect$ but it is going to inflate the variance in case of strong correlations.  

In case we consider $\mathrm{KL}[q(\wvect) || p(\wvect | X, \yvect)]$, the expression of the KL becomes:
$$
\mathrm{KL}[q(\wvect) || p(\wvect | X, \yvect)] = 
\frac{1}{2} \Tr(\Sigma^{-1} \diag(\svect^2))
+ \frac{1}{2} (\mvect - \muvect)^{\top} \Sigma^{-1} (\mvect - \muvect)
- \frac{D}{2} 
+ \frac{1}{2} \log\left( \frac{\mathrm{det}\Sigma}{\prod_i s_i^2} \right)
$$
Again, the optimal mean $\mvect$ is $\muvect$. 
For the variances $\svect^2$, we need to take the derivative of the KL divergence and set it to zero:
$$
\frac{\partial \mathrm{KL}[q(\wvect) || p(\wvect | X, \yvect)]}{\partial s^2_i} = 
 \frac{1}{2} \frac{\partial \Tr(\Sigma^{-1} \diag(\svect^2))}{\partial s_i^2}
- \frac{1}{2} \frac{\partial \sum_i \log{s_i^2}}{\partial s_i^2}
= 0
$$
Rewriting the trace term as the sum of the Hadamrd product of the matrices in the product $\sum_{ij} (\Sigma^{-1} \odot \diag(\svect^2))_{ij} = \sum_i s_i^2 \Sigma^{-1}_{ii} $, this yields 
$$
\frac{\partial \mathrm{KL}[q(\wvect) || p(\wvect | X, \yvect)]}{\partial s^2_i} = 
 \frac{1}{2} \frac{\partial s_i^2 \Sigma^{-1}_{ii}  )}{\partial s_i^2}
- \frac{1}{2} \frac{\partial \log{s_i^2}}{\partial s_i^2}
= 0
$$
This results in $(s_i^2)^{-1} = \Sigma^{-1}_{ii}$. 
This approximation has the opposite effect of underestimating the variance for each variable.

 \section{Extended results}
 \label{appex:extended-results}
\subsection{Toy example}
\begin{figure*}[!h]
    \centering
    \setlength\figureheight{.18\textwidth}
    \setlength\figurewidth{.22\textwidth} 
    \subfigure{
        \tiny
        \ifdefined\compilefigures
            \tikzsetnextfilename{demo-init-blm}
            \input{../fig/demo-1d/demo-init-blm7}
        \else
            \includegraphics{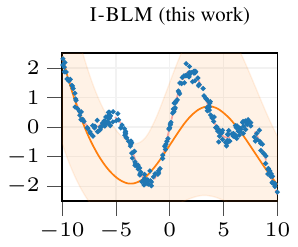}
        \fi
    }    
    \hspace{-0.04\textwidth}
    \subfigure{
        \tiny
        \ifdefined\compilefigures
        \tikzsetnextfilename{demo-init-uninf}
        \input{../fig/demo-1d/demo-init-uninf}
        \else
        \includegraphics{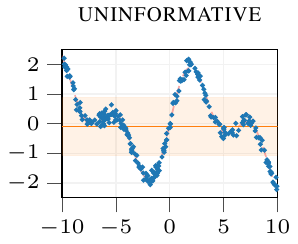}
        \fi
    }
    \hspace{-0.04\textwidth}
    \subfigure{
        \tiny
        \ifdefined\compilefigures
        \tikzsetnextfilename{demo-init-heuristic}
        \input{../fig/demo-1d/demo-init-heuristic}
        \else
        \includegraphics{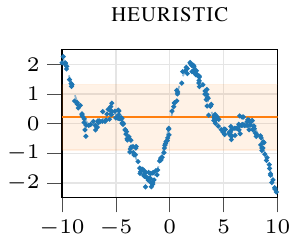}
        \fi
    }
    \hspace{-0.04\textwidth}
    \subfigure{
        \tiny
        \ifdefined\compilefigures
        \tikzsetnextfilename{demo-init-xavier}
        \input{../fig/demo-1d/demo-init-xavier}
        \else
        \includegraphics{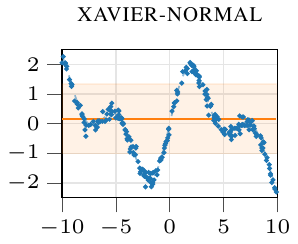}
        \fi
    }
    \hspace{-0.04\textwidth}
    \subfigure{
        \tiny
        \ifdefined\compilefigures
        \tikzsetnextfilename{demo-init-orthogonal}
        \input{../fig/demo-1d/demo-init-orth}
        \else
        \includegraphics{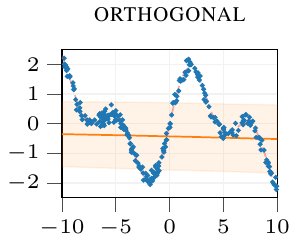}
        \fi
    }
    \hspace{-0.04\textwidth}
    \subfigure{
        \tiny
        \ifdefined\compilefigures
        \tikzsetnextfilename{demo-init-lsuv}
        \input{../fig/demo-1d/demo-init-lsuv}
        \else
        \includegraphics{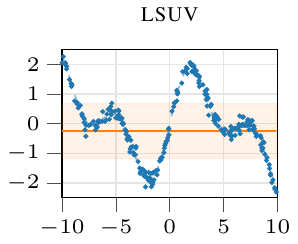}
        \fi
    }
    \caption{Predictions after initialization using our proposal and all the other competitive methods.}
    \label{fig:demo}
\end{figure*}
With this simple example we want once more illustrate how \iblm works and how it can speed up the convergence of \svi. 
We set up a regression problem considering the function $f(x) = \sin(x) + \sin(x/2) + \sin(x/3) - \sin(x/4)$ corrupted by noise $\varepsilon \sim \N(0, \exp(-2))$, with $x$ sampled uniformly in the interval $[-10,10]$. 
\autoref{fig:demo} reports the output of a $4$-layer \dnn after different initializations. 
The figure shows that \iblm obtains a sensible initialization compared to the competitors.

\subsection{Regression with shallow architecture}
\label{ssub:regression-shallow}
\begin{minipage}{\textwidth}
    
    \scriptsize
    \centering
    \begin{tabular}{@{}rcccccc@{}}
        \multicolumn{7}{c}{\normalsize \textsc{test} \rmse}\\
        \toprule
        {} & \iblm & \uninformative & \heuristic & \xavier & \orthogonal & \lsuv \\
       
        \midrule
        \powerplant & $\mathbf{0.2427 \pm 0.006}$ & $0.2452 \pm 0.007$ &
                      $0.2436 \pm 0.008$ & $0.2427 \pm 0.008$ &
                      $0.2439 \pm 0.008$ & $0.2438 \pm 0.007$ \\

        \protein & $\mathbf{0.6831 \pm 0.004}$ & $0.7135 \pm 0.008$ &
                   $0.7020 \pm 0.009$ & $0.6952 \pm 0.011$ &
                   $0.7315 \pm 0.016$ & $0.7356 \pm 0.006$ \\
        
       
        \bottomrule\\ \\
  
        \multicolumn{7}{c}{\normalsize \textsc{test} \mnll}\\
        \toprule
        {} & \iblm & \uninformative & \heuristic & \xavier & \orthogonal & \lsuv \\
       
        \midrule
        \powerplant  & $\mathbf{-0.7647 \pm 0.012}$ & $-0.7607 \pm 0.013$ &
                       $-0.7622 \pm 0.014$ & $-0.7641 \pm 0.013$ &
                       $-0.7623 \pm 0.013$ & $-0.7623 \pm 0.012$ \\

        \protein  & $\mathbf{0.7510 \pm 0.021}$ & $0.8980 \pm 0.040$ &
                    $0.8376 \pm 0.046$ & $0.8047 \pm 0.055$ &
                    $0.9878 \pm 0.083$ & $1.0081 \pm 0.033$ \\
        
         
        \bottomrule
    \end{tabular}
    \vspace{1mm}
\end{minipage}

\subsection{Regression with deep architecture}
\begin{minipage}{\textwidth}
    
    \scriptsize
    \centering
    \begin{tabular}{@{}rcccccc@{}}
        \multicolumn{7}{c}{\normalsize \textsc{test} \rmse}\\
        \toprule
        {} & \iblm & \uninformative & \heuristic & \xavier & \orthogonal & \lsuv \\
       
        \midrule
        \powerplant &
            $0.2472 \pm 0.003$ & $0.2476 \pm 0.005$ &
            $\mathbf{0.2462 \pm 0.005}$ & $0.2658 \pm 0.030$ &
            $0.2467 \pm 0.005$ & $0.2774 \pm 0.026$ \\
        
        \protein & 
            $\mathbf{0.6683 \pm 0.007}$ & $0.7170 \pm 0.013$ & 
            $0.6899 \pm 0.011$ & $0.6821 \pm 0.007$ & 
            $0.6982 \pm 0.014$ & $0.7033 \pm 0.011$ \\
                
        \bottomrule\\ \\
  
        \multicolumn{7}{c}{\normalsize \textsc{test} \mnll}\\
        \toprule
        {} & \iblm & \uninformative & \heuristic & \xavier & \orthogonal & \lsuv \\
        \midrule
        \powerplant  & 
            $\mathbf{-0.7455 \pm 0.008}$ & $-0.7420 \pm 0.008$ & 
            $-0.7455 \pm 0.009$ & $-0.7007 \pm 0.070$ & 
            $-0.7450 \pm 0.010$ & $-0.6677 \pm 0.065$ \\

        \protein  & 
            $\mathbf{0.6922 \pm 0.035}$ & $0.9326 \pm 0.066$ & 
            $0.7884 \pm 0.055$ & $0.7540 \pm 0.033$ & 
            $0.8280 \pm 0.072$ & $0.8587 \pm 0.040$ \\
        
           
        \bottomrule
    \end{tabular}
    \vspace{1mm}
\end{minipage}

\subsection{Classification with shallow architecture}
\begin{figure}[!h]
    \centering
    \setlength\figureheight{.19\textwidth}
    \setlength\figurewidth{.28\textwidth} 
    \hspace{-0.045\textwidth}
    \subfigure{
        \tiny
        \ifdefined\compilefigures
        \tikzsetnextfilename{classification-shallow-mnist}
        \input{../fig/class-exp/test_5folds_mnist_6methods.tex}
        \else
        \includegraphics{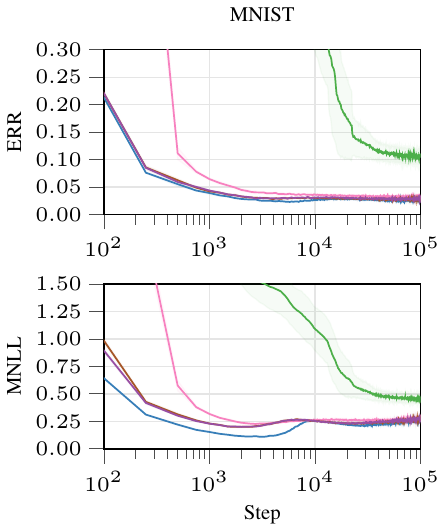}
        \fi
    }
    \hspace{-0.025\textwidth}
    \subfigure{
        \tiny
        \ifdefined\compilefigures
        \tikzsetnextfilename{classification-shallow-spam}
        \input{../fig/class-exp/test_5folds_spam_6methods.tex}
        \else
        \includegraphics{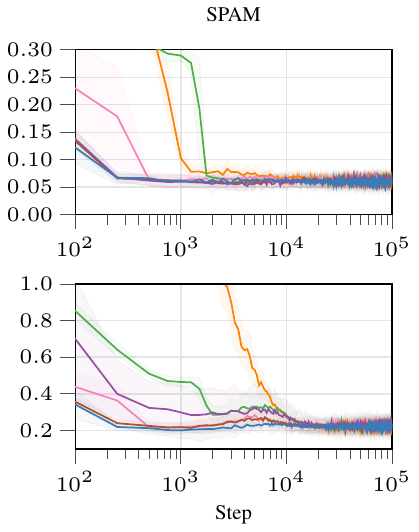}
        \fi
    }
    \hspace{-0.025\textwidth}
    \subfigure{
        \tiny
        \ifdefined\compilefigures
        \tikzsetnextfilename{classification-shallow-eeg}
        \input{../fig/class-exp/test_5folds_eeg_6methods.tex}
        \else
        \includegraphics{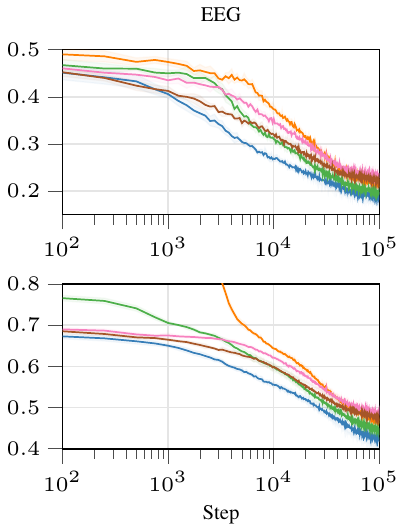}
        \fi
    }
    \hspace{-0.025\textwidth}
    \subfigure{
        \tiny
        \ifdefined\compilefigures
        \tikzsetnextfilename{classification-shallow-credit}
        \input{../fig/class-exp/test_5folds_credit_6methods.tex}
        \else
        \includegraphics{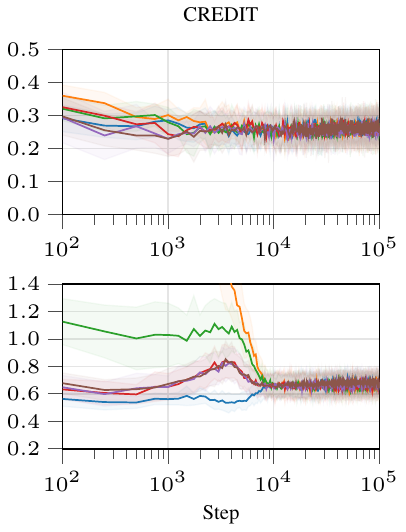}
        \fi
    }
    \\[-1ex]
    \subfigure{
        \tiny
        \ifdefined\compilefigures
        \tikzsetnextfilename{classification-shallow-legend}
        \input{../fig/class-exp/legend_6methods}
        \else
        \includegraphics{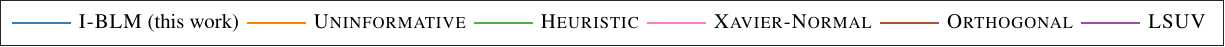}
        \fi
    }
    \caption{Progression of test \errate and test \mnll over training iterations for different initialization strategies on four classification datasets.}


\end{figure}

\begin{minipage}{\textwidth}
    
    \scriptsize
    \centering
    \begin{tabular}{@{}rcccccc@{}}
        \multicolumn{7}{c}{\normalsize \textsc{test} \errate}\\
        \toprule
        {} & \iblm & \uninformative & \heuristic & \xavier & \orthogonal & \lsuv \\
       
        \midrule
        \spam & $\mathbf{0.0594 \pm 0.013}$ & $0.0620 \pm 0.008$ 
             & $0.0624 \pm 0.011$ & $0.0620 \pm 0.012$ 
             & $0.0611 \pm 0.012$ & $0.0598 \pm 0.014$\\

        \eeg & $\mathbf{0.1855 \pm 0.015} $ & $0.1929 \pm 0.009 $ 
            & $0.2221 \pm 0.009 $ & $0.2137 \pm 0.008 $ 
            & $0.2335 \pm 0.007 $ & NC\\
        
        \credit & $0.2680 \pm 0.027 $ & $0.2679 \pm 0.044 $ 
               & $\mathbf{0.2480 \pm 0.071} $ & $0.2519 \pm 0.033 $ 
               & $0.2539 \pm 0.032 $ & $0.2580 \pm 0.050 $ \\

        \mnist & $\mathbf{0.0253 \pm 7\mathrm{e}{-4}}$ & NC 
              & $0.1046 \pm 0.014$ & $0.0315 \pm 0.001$ 
              & $0.0275 \pm 0.001$ & $0.0291 \pm 0.002$  \\
       
        \bottomrule\\ \\
  
        \multicolumn{7}{c}{\normalsize \textsc{test} \mnll}\\
        \toprule
        {} & \iblm & \uninformative & \heuristic & \xavier & \orthogonal & \lsuv \\
       
        \midrule
        \spam & $0.229 \pm 0.034$ & $\mathbf{0.213 \pm 0.030}$ 
             & $0.228 \pm 0.043$ & $0.228 \pm 0.048$ 
             & $0.225 \pm 0.053$ & $0.228 \pm 0.050$ \\

        \eeg & $\mathbf{ 0.4218 \pm 0.020}$ & $0.4668 \pm 0.008$ 
            & $0.4411 \pm 0.006$ & $0.4866 \pm 0.010$ 
            & $0.4728 \pm 0.010$ & NC \\
        
        \credit & $0.6759 \pm 0.084$ & $\mathbf{0.6597 \pm 0.101}$ 
               & $0.6605 \pm 0.111$ & $0.6616 \pm 0.105$ 
               & $0.6662 \pm 0.076$ & $0.6739 \pm 0.069$ \\

        \mnist & $0.2655 \pm 0.015$ & NC 
              & $0.4497 \pm 0.039$ & $0.2724 \pm 0.020$ 
              & $\mathbf{0.2643 \pm 0.017}$ & $0.2744 \pm 0.014$ \\

        \bottomrule
    \end{tabular}
    \vspace{1mm}
\end{minipage}    

\subsection{Classification with deep architecture}
\begin{minipage}{\textwidth}
    
    \scriptsize
    \centering
    \begin{tabular}{@{}rcccccc@{}}
        \multicolumn{7}{c}{\normalsize \textsc{test} \errate}\\
        \toprule
        {} & \iblm & \uninformative & \heuristic & \xavier & \orthogonal & \lsuv \\
       
        \midrule
        \mnist &
            $\mathbf{0.0356 \pm 0.003}$ & $0.0390 \pm 0.003$ & 
            $0.0400 \pm 0.003$ & $0.0411 \pm 0.002$ & 
            $0.0396 \pm 0.002$ & $0.0373 \pm 0.001$ \\
        
        \eeg & 
            $\mathbf{0.0673 \pm 0.008}$ & $0.1283 \pm 0.009$ &
            $0.1119 \pm 0.008$ & $0.0894 \pm 0.003$ &
            $0.1216 \pm 0.002$ & NC \\

        \credit & 
            $\mathbf{0.2700 \pm 0.024}$ & $0.2975 \pm 0.059$ & 
            $0.2824 \pm 0.058$ & $0.2833 \pm 0.022$ & 
            $0.3145 \pm 0.051$ & $0.2758 \pm 0.022$ \\

        \spam & 
            $0.0566 \pm 0.021$ & $0.0611 \pm 0.008$ & 
            $0.0585 \pm 0.017$ & $0.0534 \pm 0.018$ & 
            $\mathbf{0.0514 \pm 0.013}$ & $0.0611 \pm 0.013$ \\

        \bottomrule\\ \\
  
        \multicolumn{7}{c}{\normalsize \textsc{test} \mnll}\\
        \toprule
        {} & \iblm & \uninformative & \heuristic & \xavier & \orthogonal & \lsuv \\
       
        \midrule
        \mnist  & 
            $\mathbf{0.1692 \pm 0.007}$ & $0.1847 \pm 0.002$ & 
            $0.1799 \pm 0.009$ & $0.1912 \pm 0.011$ & 
            $0.1822 \pm 0.005$ & $0.1723 \pm 0.005$ \\

        \eeg & 
            $\mathbf{0.4222 \pm 0.054}$ & $1.2515 \pm 0.352$ &
            $0.8136 \pm 0.123$ & $0.6273 \pm 0.130$ &
            $0.9366 \pm 0.097$ & NC\\

        \credit & 
            $2.6555 \pm 0.521$ & $3.2836 \pm 0.704$ &
            $3.1268 \pm 0.678$ & $2.7015 \pm 0.665$ &
            $2.6482 \pm 0.231$ & $\mathbf{2.5422 \pm 0.236}$ \\

        \spam & 
            $\mathbf{0.7021 \pm 0.218}$ & $1.1098 \pm 0.271$ &
            $1.0458 \pm 0.517$ & $1.0682 \pm 0.347$ &   
            $0.8176 \pm 0.337$ & $1.1682 \pm 0.486$\\
           
        \bottomrule
    \end{tabular}
\end{minipage}    

\subsection{Convolutional neural networks}
\begin{minipage}{\textwidth}
    \centering
    \scriptsize
    \begin{tabular}{@{}rcccccc@{}}
        \multicolumn{7}{c}{\normalsize \textsc{test} \errate}\\
        \toprule
        {} & \iblm & \uninformative & \heuristic & \xavier & \orthogonal & \lsuv \\
       
        \midrule
        \mnist &
            $\mathbf{0.0087}$ & NC & NC & NC & $0.0098$ & $0.0113$ \\
        
        \cifart & 
            $\mathbf{0.3499 }$ & NC & NC & NC & $0.3784 $ & $0.3846 $  \\

        \bottomrule\\ 
  
        \multicolumn{7}{c}{\normalsize \textsc{test} \mnll}\\
        \toprule
        {} & \iblm & \uninformative & \heuristic & \xavier & \orthogonal & \lsuv \\
       
        \midrule
        \mnist  & 
            $\mathbf{0.0345 }$ & NC & NC & NC & $0.0377 $ & $0.0421 $ \\

        \cifart & 
            $\mathbf{1.0683 }$ & NC & NC & NC & $1.1270 $ & $1.1428 $\\

        \bottomrule
    \end{tabular}
    \vspace{1mm}
    \label{tab:cnns-exp}
\end{minipage}

\subsection{Uncertainty estimation}
\begin{figure}[!h]
    \centering
    \hspace{-0.062\textwidth}
    \subfigure{
        \setlength\figureheight{.25\textwidth}
        \setlength\figurewidth{.35\textwidth} 
        \tiny
        \ifdefined\compilefigures
        \tikzsetnextfilename{lenet_entropy}
        \input{../fig/montecarlo-exp/entropy}
        \else
        \includegraphics{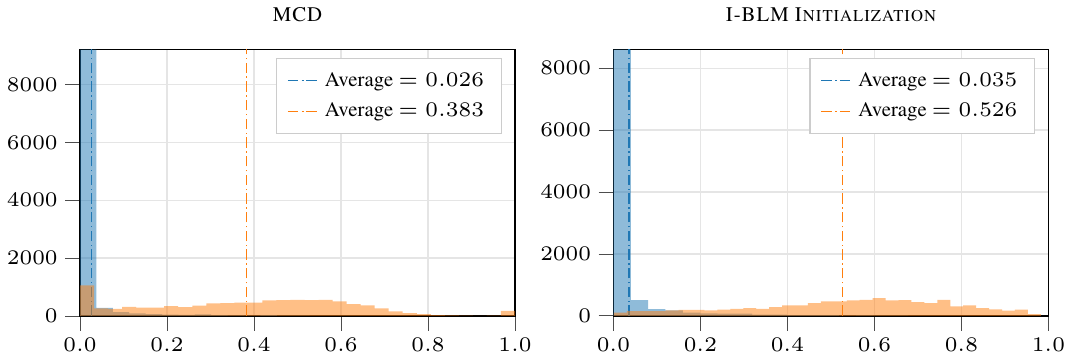}
        \fi
    } 
    \\[-1ex]
    \subfigure{
        \tiny
        \ifdefined\compilefigures
        \tikzsetnextfilename{lenet_entropy_legend}
        \input{../fig/montecarlo-exp/legend-entropy}
        \else
        \includegraphics{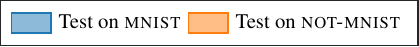}
        \fi
    }
    \caption{Entropy distribution while testing on \mnist and \notmnist (higher average entropy on \notmnist means better uncertainty estimation). }
    \label{fig:entropy}
\end{figure}


One of the advantages of Bayesian inference is the possibility to reason about uncertainty. 
With this experiment, we aim to demonstrate that \svi with a Gaussian approximate posterior is competitive with \mcd in capturing uncertainty in predictions. 
To show this, we focus on a \cnn with the \lenet architecture. 
We run \mcd and \svi with a Gaussian approximate posterior with the proposed initialization on \mnist.
At test time, we carry out predictions on both \mnist and \notmnist; the latter is a dataset equivalent to \mnist in input dimensions ($1\times28\times28$) and number of classes, but it represents letters rather than numbers\footnote{\url{http://yaroslavvb.blogspot.com/2011/09/notmnist-dataset.html}}.
This experimental setup is often used to check that the entropy of the predictions on \notmnist are actually higher than the entropy of the predictions on \mnist. 
We report the entropy of the prediction on \mnist and \notmnist in \autoref{fig:entropy}. 
\mcd and \svi behave similarly on \mnist, but on \notmnist the the histogram of the entropy indicates that \svi yields a slightly higher uncertainty compared to \mcd. 

\subsection{Calibration}

\begin{figure}[!h]
    \centering
    \hspace{-0.03\textwidth} 
    \setlength\figureheight{.25\textwidth}
    \setlength\figurewidth{.35\textwidth} 
    \subfigure{
        \tiny
        \ifdefined\compilefigures
        \tikzsetnextfilename{calibration_mnist}
        \input{../fig/calibration-exp/calibration-mnist}
        \else
        \includegraphics{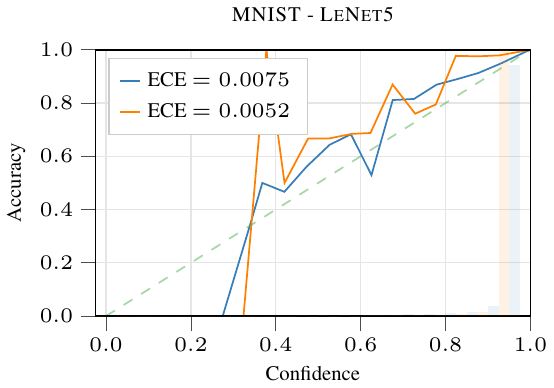}
        \fi
    }
    \hspace{-0.03\textwidth}
    \subfigure{
        \tiny
        \ifdefined\compilefigures
        \tikzsetnextfilename{calibration_cifar10}
        \input{../fig/calibration-exp/calibration-cifar10}
        \else
        \includegraphics{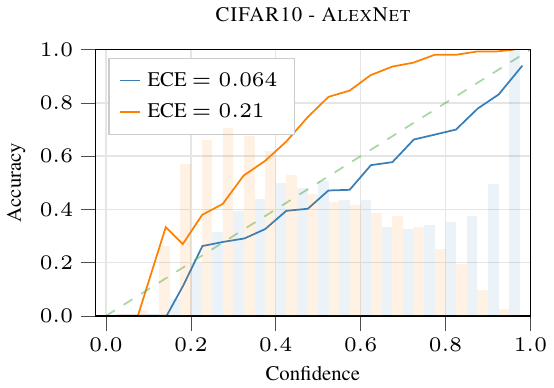}
        \fi
    }
    \\[-1ex]
    \subfigure{
        \tiny
        \ifdefined\compilefigures
        \tikzsetnextfilename{calibration_legend}
        \input{../fig/calibration-exp/legend}
        \else
        \includegraphics{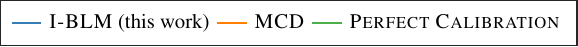}
        \fi
    }
    \caption{Comparison of reliability diagrams and \ece  between \iblm and \mcd on \mnist (\textbf{left}) and \cifart (\textbf{right}).}
    \label{fig:calibration-exp}
\end{figure}

Calibration of uncertainty is an important performance metric that one should take into account for comparing classification models \citep{Flach16, Guo17}. 
\textit{Reliability Diagrams} and the \textit{Expected Calibration Error} are standard methods to empirically estimate the calibration uncertainty.
\textit{Reliability Diagrams} are a visualization tool where sample accuracy is plotted as function of confidence \citep{DeGroot1983, Niculescu-Mizil05}. 
For a perfectly calibrated model, the diagram follows the identity function. 
\textit{Expected Calibration Error} (or \ece) represents a summary statistic of the calibration \citep{Naeini15}.
\autoref{fig:calibration-exp} shows the reliability diagrams and the \ece for \lenet trained on \mnist and for \alexnet trained on \cifart. 
Even tough they show similar properties on \mnist, with \alexnet on \cifart, \svi initialized with \iblm improves the calibration of uncertainty up to $3.5$ times over \mcd. 

\subsection{\KL regularization policy for Gaussian \svi}
The \KL regularization term in the variational objective severely penalizes training of over-parameterized model. 
With a sensible initialization of this kind of model, the approximate posterior is drastically different from a spherical Gaussian prior and the variational objective is majorly dominated by the regularization term rather than the reconstruction likelihood. 
To deal with such issue, we propose and implement a simple policy to gradually include the \KL term in the \nelbo.
Given the generic expression for the \nelbo, we modify the lower bound as follow:
\begin{equation*} 
    \nelbo = \nll + \lambda \dkl{q_{\thetavect}(\Wvect)}{p(\Wvect)} 
    \quad \text{where} \quad
    \lambda = \gamma\left( 1 + \exp(-\alpha(\text{iter} - \beta)) \right)^{-1} \, , 
\end{equation*}
This way, we start the optimization of the \nelbo with low regularization, and progressively increase it throughout the optimization. 
For the experiment on \vgg, we used $\alpha=2\cdot10^{-3}$, $\beta=2.5 \cdot 10^{4}$ and $\gamma=10^{-1}$.

\end{document}